\documentclass[11pt]{article}

\usepackage[preprint]{include/acl}

\usepackage{times}
\usepackage{latexsym}

\usepackage[T1]{fontenc}

\usepackage[utf8]{inputenc}

\usepackage{microtype}

\usepackage{inconsolata}

\usepackage{graphicx}

\usepackage{hyperref}
\usepackage{url}
\usepackage{booktabs}
\usepackage{multirow} 

\definecolor{darkblue}{rgb}{0, 0, 0.5}
\hypersetup{colorlinks=true, citecolor=darkblue, linkcolor=darkblue, urlcolor=darkblue}

\usepackage{algorithm}
\usepackage{algorithmic}

\usepackage{amsmath}
\usepackage{amssymb}
\usepackage{mathtools}
\usepackage{amsthm}

\usepackage{enumitem}
\usepackage{xcolor} 
\usepackage{soul}   

\usepackage[capitalize,noabbrev]{cleveref}

\theoremstyle{plain}
\newtheorem{theorem}{Theorem}[section]

\newtheorem{corollary}[theorem]{Corollary}
\theoremstyle{definition}

\newtheorem{example}[theorem]{Example}
\theoremstyle{remark}

\usepackage[textsize=tiny]{todonotes}

\usepackage[most]{tcolorbox}

\newtcolorbox{promptbox}[1][]{
  colback=gray!5!white,      
  colframe=gray!75!black,    
  coltitle=white,            
  fonttitle=\bfseries,       
  fontupper=\ttfamily,       
  rounded corners,
  arc=4mm,                   
  boxrule=1pt,               
  left=10pt, right=10pt, top=10pt, bottom=10pt, 
  breakable,                 
  enhanced,                  
  #1                         
}

\usepackage{amsmath,amsfonts,bm}

\def\eqref#1{equation~\ref{#1}}

\def\1{\bm{1}}

\DeclareMathAlphabet{\mathsfit}{\encodingdefault}{\sfdefault}{m}{sl}
\SetMathAlphabet{\mathsfit}{bold}{\encodingdefault}{\sfdefault}{bx}{n}

\newcommand{\KL}{D_{\mathrm{KL}}}

\newcommand{\Expectation}{\ensuremath{\mathbb{E}}}

\newcommand{\user}{\ensuremath{p}}
\newcommand{\prompt}{\ensuremath{q}}

\newcommand{\refpolicy}{\ensuremath{\pi_{\text{ref}}}}
\newcommand{\policy}{\ensuremath{\pi_\theta}}

\newcommand{\reward}{\ensuremath{R}}
\newcommand{\advantage}{\ensuremath{\hat{A}}}

\title{Personalized Group Relative Policy Optimization for Heterogenous Preference Alignment}

\author{
  \textbf{Jialu Wang\textsuperscript{1}},
  \textbf{Heinrich Peters\textsuperscript{1}},
  \textbf{Asad A. Butt\textsuperscript{1}},
  \textbf{Navid Hashemi\textsuperscript{1}},
  \textbf{Alireza Hashemi\textsuperscript{1}},
\\
  \textbf{Pouya M. Ghari\textsuperscript{1}},
  \textbf{Joseph Hoover\textsuperscript{1}},
  \textbf{James Rae\textsuperscript{1}},
  \textbf{Morteza Dehghani\textsuperscript{1}}
\\
\\
  \textsuperscript{1}Apple Inc. \\
}

\begin{document}

\maketitle

\begin{abstract}
    Despite their sophisticated general-purpose capabilities, Large Language Models (LLMs) often fail to align with diverse individual preferences because standard post-training methods, like Reinforcement Learning with Human Feedback (RLHF), optimize for a single, global objective. While Group Relative Policy Optimization (GRPO) is a widely adopted on-policy reinforcement learning framework, its group-based normalization implicitly assumes that all samples are exchangeable. This assumption conflates distinct user reward distributions and systematically biases learning toward dominant preferences while suppressing minority signals. To address this problem, we introduce \emph{Personalized GRPO} (P-GRPO), a novel alignment framework that decouples advantage estimation from immediate batch statistics. By normalizing advantages against preference-group-specific reward histories rather than the concurrent generation group, P-GRPO preserves the contrastive signal necessary for learning distinct preferences. We evaluate P-GRPO across diverse tasks and find that it consistently achieves faster convergence and higher rewards than standard GRPO, thereby enhancing its ability to recover and align with heterogeneous preference signals. Our results demonstrate that accounting for reward heterogeneity at the optimization level can be helpful for building models that faithfully align with diverse human preferences without sacrificing general capabilities.

\end{abstract}

\section{Introduction}

Large Language Models (LLMs) have demonstrated remarkable performance across a wide range of general-purpose tasks, including reasoning, dialogue, and content recommendation \citep{radford2019language,johnsen2024large}. However, as these models are increasingly deployed in interactive and high-stakes settings, ensuring that their behaviors are aligned with human preferences and values has become a central challenge \citep{Shen2023LargeLM}. Modern alignment approaches typically rely on post-training procedures such as supervised fine-tuning \citep{radford2018improving} and Reinforcement Learning from Human Feedback (RLHF), which optimize the model with respect to a learned reward signal intended to reflect desirable behavior \citep{ouyang2022training, wang2024reinforcementlearningenhancedllms}. Implicit in most existing alignment approaches is the assumption that this reward function represents a \emph{single, homogeneous preference signal} that is broadly shared across users and contexts \citep{park2024rlhfheterogeneousfeedbackpersonalization}.

In practice, however, human preferences are neither homogeneous \citep{zhang2025diverging,10.1145/3757887.3767678} nor stationary \citep{zafari2018modellinganalysistemporalpreference,son2025right}. Preferences vary systematically across individuals, cultures, psychological dispositions, and situational contexts. For example, different users may prefer concise versus detailed explanations, neutral versus emotionally expressive language, or exploratory versus conservative recommendations. Such differences are well documented across cultural communication styles \citep{hofstede1984hofstede,holtgraves1997styles}, personality traits  \citep{mccrae1987validation,10.1093/pnasnexus/pgae231,matz2024potential}, and behavioral patterns in interactive systems \citep{pirolli1999information,rose2004understanding,shani2010evaluating}. From a modeling perspective, this implies that the reward signal observed during alignment is better viewed as arising from a mixture of user- or context-conditioned preference functions rather than from a single global objective. Alignment methods that ignore this heterogeneity risk systematically optimizing for dominant preference modes while degrading performance for under-represented or minority patterns \citep{sourati2025shrinkinglandscapelinguisticdiversity}.

Reinforcement learning (RL) has emerged as a particularly effective paradigm for post-training alignment, as it allows models to explore a large space of possible outputs and optimize directly for preference-based reward signals \citep{deepseekai2025deepseekr1incentivizingreasoningcapability}. Among recent approaches, Group Relative Policy Optimization (GRPO) and its variants have become a dominant framework for stable on-policy optimization \citep{deepseek-math,yu2025dapo,gspo,hu2025reinforcestabilizingcriticfreepolicy}. GRPO operates by sampling a group of completion trajectories for each prompt, evaluating them using a reward model or verifiable outcome, and computing token-level advantages by normalizing each trajectory’s reward relative to others within the same group. This group-based normalization reduces variance and stabilizes training, but it implicitly assumes that all rewards within a group are exchangeable samples from the same underlying preference distribution.

\begin{figure*}[t]
    \centering
    \includegraphics[width=0.9\linewidth]{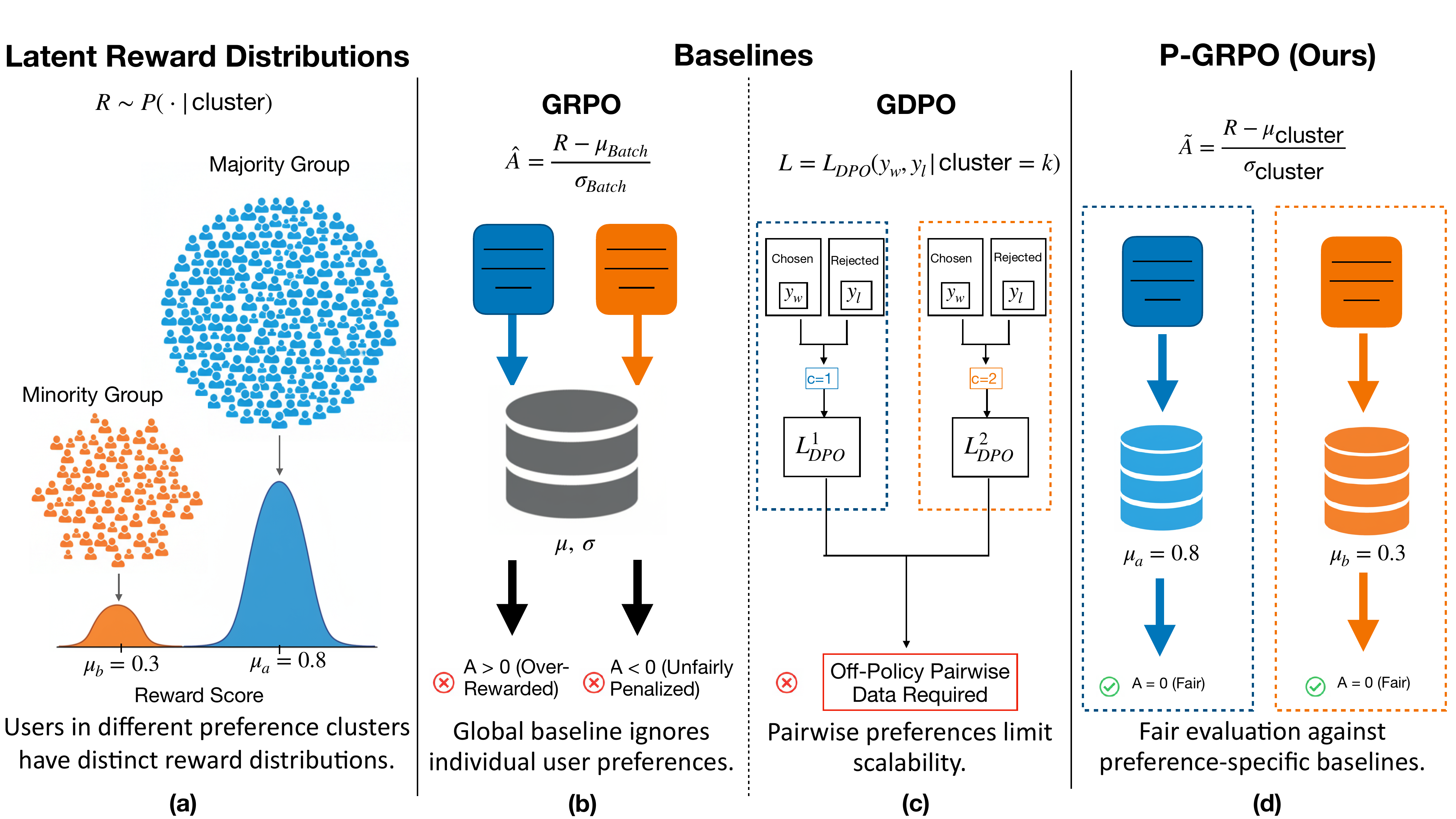}
    \caption{\textbf{Overview of Personalized Group Relative Policy Optimization (P-GRPO).} 
  \textbf{(a) Latent Reward Distributions:} Users in different preference clusters often exhibit distinct reward distributions. In this example, a majority group (Blue) has a high mean reward ($\mu \approx 0.8$), while a minority group (Orange) has a lower mean reward ($\mu \approx 0.3$).
  \textbf{(b) GRPO:} Standard GRPO normalizes rewards using a generation batch mean ($\mu_\text{Batch}$). 
  \textbf{(c) GDPO:} GDPO~\citep{yao2025no} conditions DPO loss on cluster membership, requiring pairwise preference data (chosen $y_w$ vs.\ rejected $y_l$) for each cluster. 
  \textbf{(d) P-GRPO (Ours):} Our method normalizes rewards against preference-specific statistics ($\mu_{\text{cluster}}, \sigma_{\text{cluster}}$). 
  By comparing each output against its own cluster-wise baseline (e.g., $\mu_b \approx 0.3$ for minority users), P-GRPO correctly assigns advantages ($\tilde{A} \approx 0$), ensuring equitable optimization across diverse user preferences.
  }

    \label{fig:p-grpo}
\end{figure*}

This assumption breaks down in the presence of heterogeneous preferences. When rewards reflect a mixture of distinct preference functions, group-based normalization induces a form of statistical shrinkage toward the dominant preference mode: trajectories associated with preference modes that are rarer, noisier, or systematically lower-variance contribute weaker or noisier gradients after normalization. As a result, standard GRPO can converge to policies that perform well for the most common preferences in the training data while under-performing for other users or contexts. This dynamic limits the ability of current alignment methods to support personalization and can lead to systematic disparities in model quality across user groups.

In this work, we propose \emph{Personalized Group Relative Policy Optimization (P-GRPO)}, a modification of GRPO designed to learn distinct user preferences without collapsing toward a single majority opinion. By normalizing rewards against the historical statistics of a user's specific preference cluster, P-GRPO preserves the rich, contrastive signals in heterogeneous reward data, which are often lost in standard group normalization. While we focus on personalization settings in which preference structure is either observed explicitly (e.g., via user identifiers) or recoverable through clustering of interaction signals, the method applies more generally to any setting with structured reward heterogeneity. Figure~\ref{fig:p-grpo} provides an overview of our method.

We evaluate P-GRPO on several controlled tasks that instantiate heterogeneous reward settings: (1) a content recommendation task, (2) a set of preference-conditioned text generation tasks. Experiments are conducted using Qwen3 \citep{qwen3} and Gemma \citep{gemmateam2024gemma2improvingopen} models at multiple scales.
Across all tasks and settings, P-GRPO outperforms standard GRPO in both convergence speed and average reward. LLM-as-judge evaluations confirm these gains, with higher win rates indicating superior alignment with diverse user objectives. Furthermore, we show in Appendix~\ref{apdx/mmlu} that these personalization gains do not compromise general reasoning capabilities, as P-GRPO maintains robust performance on the MMLU benchmark~\citep{hendrycks2021measuring}. 
\section{Related Work}
Personalization in LLMs aims to adapt general-purpose systems to individual users’ preferences, goals, or contexts \citep{zhang2025personalization, guan-etal-2025-survey}. Existing approaches can be broadly grouped into three categories: inference-time personalization, representation-based personalization, and optimization-level personalization.

Inference-time personalization methods modify the input context to steer model behavior without changing the model parameters. This includes approaches such as personalized prompting \citep{jiang2023evaluating}, injecting explicit user profiles into prompts \citep{dipalma2024evaluatingchatgptrecommendersystem}, and retrieval-augmented generation (RAG) methods that incorporate user summaries or behavioral histories \citep{richardson2023integratingsummarizationretrievalenhanced}. These methods are flexible and low-cost but rely on the model’s pre-existing capacity to respond appropriately to conditioning signals, and they do not directly address how models are trained to represent and optimize for diverse preferences. Another line of work considers leveraging reward-guided decoding to provide personalization \citep{bu-etal-2025-personalized}, but requires a reward model that can accurately capture the user preferences at test time.

Representation-based approaches learn user or preference embeddings that are incorporated into the model’s input or internal state. This paradigm, demonstrated in large-scale recommender systems \citep{covington2016deep}, has been adapted for modern LLMs. Recent works, for instance, learn user-specific representations from interaction data to steer model generation towards a user's preferences \citep{ren2024representation,10.1145/3701716.3715463}. While effective for capturing stable user traits, these approaches still rely on a shared global training objective and do not address how heterogeneous preference signals shape learning dynamics.

Optimization-level personalization methods modify the learning objective itself to account for preference diversity. Personalized Soups \citep{jang2023personalized} formulate personalization as a multi-objective reinforcement learning problem, while Personalized-RLHF \citep{personalizedRLHF} jointly learns a user model and a personalized reward function. Rewards-in-Context \citep{yang2024rewards} proposes to encode the heterogeneous rewards explicitly in the prompt context and aligns with SFT. Variational Preference Learning \citep{poddar2024personalizing} introduces latent preference variables inferred via variational encoders, and Group Distributional Preference Optimization \citep[GDPO;][]{yao2025no} conditions preference optimization on group membership. FSPO \citep{singh2025fspofewshotpreferenceoptimization} frames personalization as a meta-learning problem to enable rapid adaptation with few examples.

Our work belongs to this third category but differs in its focus and paradigm. Rather than introducing new preference representations or conditioning mechanisms, we study the dynamics of on-policy reinforcement learning under heterogeneous reward distributions. Unlike prior optimization procedures, we propose a modification to the on-policy optimization objective that prevents systematic suppression of minority preference signals.

\section{Approach}


\subsection{Preliminary}\label{sec:approach/preliminary}
GRPO is a policy gradient method that optimizes language models by comparing multiple sampled completions for the same prompt, using their relative quality to guide learning. This in-group comparison stabilizes training by reducing the variance in reward signals~\citep{deepseek-math}.

Let $\user$ represent the individual user preference and $\prompt$ represent the input prompt. For each prompt, GRPO samples a group of $G$ completions $\{o_i\}_{i=1}^G$ from a behavior policy $\refpolicy(\cdot\mid\user, \prompt)$, which is typically a frozen reference model that prevents the policy from deviating too far during training. The algorithm then updates the trainable policy $\policy$ by maximizing the following objective:
\begin{equation}\label{eq:L_GRPO}
    \mathcal{L}_{\text{GRPO}} = \Expectation\left[ \frac{1}{G} \sum_{i=1}^G \frac{1}{|o_i|} \sum_{t=1}^{|o_i|} \mathcal{J}_{i,t}(\theta) \right]
\end{equation}
where $\mathcal{J}_{i, t}$ represents the token-level loss that combines a clipped policy gradient objective with a KL regularization term:
\begin{multline}\label{eq:token-wise-GRPO}
    \mathcal{J}_{i,t}(\theta) = \min \left(\text{clip}\left(\rho_{i,t}, 1-c, 1+c\right) \advantage_{i,t}, \right. \\ 
    \left.\rho_{i,t} \advantage_{i,t} \right) - \beta \KL(\policy \| \refpolicy).
\end{multline}
The first term employs the clipped surrogate objective from PPO, which prevents excessively large policy updates by constraining the importance sampling ratio $\rho_{i,t}$ to lie within $[1-c, 1+c]$. This ratio measures how much the new policy $\policy$ differs from the reference policy $\refpolicy$ at each token:
\begin{equation}\label{eq:importance-sampling}
    \rho_{i,t} = \frac{\policy(o_{i,t} \mid \user, \prompt, o_{i, <t} ) }{\refpolicy(o_{i,t} \mid \user, \prompt, o_{i, <t} )},
\end{equation}
where $o_{i,t}$ is the $t$-th token in completion $o_i$, and $o_{i,<t}$ denotes all preceding tokens. The second term in $\mathcal{J}_{i,t}$ is the per-token KL divergence $\KL(\policy \| \refpolicy)$, weighted by hyperparameter $\beta$, which explicitly penalizes the policy for deviating too far from the reference policy.

A key component of GRPO is the advantage function $\advantage_{i,t}$, which determines whether a particular completion should be reinforced or discouraged. The advantage is computed by normalizing the scalar reward $\reward_i$, corresponding to completion $o_i$, with other completions in the same group:
\begin{equation}
    \advantage_{i,t} = \frac{\reward_i - \text{mean}(\{\reward_i\}_{i=1}^G)}{\text{std}(\{\reward_i\}_{i=1}^G) + \epsilon},
\end{equation}
where $\epsilon$ is a small constant to avoid division by zero. This group normalization strategy has two benefits: (1) it reduces variance by comparing completions generated from the same prompt under similar conditions, and (2) it provides a relative ranking signal that is invariant to the absolute scale of rewards. However, as we will show, this normalization scheme can lead to biased optimization that favors majority preferences.

\subsection{P-GRPO}\label{sec:approach/P-GRPO}
While the GRPO loss function effectively stabilizes training via group normalization, it implicitly treats reward signals across user preferences as exchangeable, thereby obscuring systematic differences across user preference regimes. In practice, user preferences exhibit inherent heterogeneity: certain preferences are easier to satisfy and universally yield higher rewards (e.g., users who prefer concise responses), whereas other preferences are harder to satisfy and consistently yield lower rewards (e.g., users who prefer highly technical, domain-specific responses). By normalizing only within the current generation group of size $G$, standard GRPO discards this crucial context and does not distinguish the rewards generated over diverse user preference distributions (see the example in Appendix~\ref{apdx/linear_reward}).

To address this limitation, we introduce \textit{Personalized GRPO} (P-GRPO), which normalizes advantages relative to preference-specific reward distributions rather than within-group distributions. Our approach is based on the assumption that the user population can be partitioned into meaningful preference groups, either through explicit user identification or through clustering of implicit preference signals. We assume that (1) different users or user clusters exhibit distinct preferences, (2) these preference groups have separable reward distributions with potentially different means and variances, and (3) preference identifiers or cluster assignments are available at training time. Under these assumptions, we can maintain separate running statistics for each preference group $p$.  

Instead of normalizing over the same generation group, we normalize the advantage relative to all historic rewards corresponding to the user preference $\user$. Specifically, we maintain running statistics for each preference group $p$, denoting the mean and standard deviation as:
\begin{align}
    \mu_p = \text{mean}(\{\reward_i \mid p\}) \qquad
    \sigma_p = \text{std}(\{\reward_i \mid p\}) ,
\end{align}
where $\{\reward_i \mid p\}$ represents the rewards of all completions observed for preference $p$. These statistics capture the typical reward level and variability for each preference group, providing a personalized baseline for evaluating new completions. We then compute the personalized advantage as:
\begin{align}\label{eq:userwise-advantage}
    \tilde{A}_{i,t}^p = \frac{\reward_i - \mu_p}{\sigma_p + \epsilon}.
\end{align}

This formulation ensures that the advantage signal reflects how well a completion performs relative to that preference group's typical performance, rather than relative to other completions in the same generation batch. By accounting for preference-specific reward biases, P-GRPO provides equitable learning signals across diverse user groups: a high-reward completion for an easy preference and a moderate-reward completion for a hard preference can both receive appropriate gradient updates that reflect their true value to their respective users. We defer the practical implementation of the algorithm to Appendix~\ref{appx:pratical-implementation}.

\begin{corollary}\label{cor:decomposition}
The cluster-level advantage $\tilde{A}_{i,t}^p$ defined in \eqref{eq:userwise-advantage} decomposes into a rescaled group advantage and a bias correction term:
\begin{equation}\label{eq:advantage-decomposition}
    \tilde{A}_{i,t}^p =  \frac{\sigma_G}{\sigma_p}\hat{A}_{i,t} + \frac{\mu_G - \mu_p}{\sigma_p},
\end{equation}
where $\hat{A}_{i,t}$ is the GRPO advantage normalized within the generation group, $\mu_G$ is the mean reward over the current generation group, and $\mu_p$ is the historical mean reward for preference group $p$. 
\end{corollary}



\section{Experiments}\label{sec:exp}

We evaluate P-GRPO on two use cases: preference-aligned content recommendation, using MovieLens-1M \citep{konstan1997grouplens}, and preference-aligned language generation, using a novel synthetic preference dataset, Goodreads Book Reviews (\emph{Goodreads}) \citep{DBLP:conf/acl/WanMNM19}, and KGRec-Music (\emph{KGRec}) \citep{kgrec-music}.

\subsection{Setup}
\subsubsection{Datasets and Training Tasks}

\paragraph{MovieLens-1M} \citep{konstan1997grouplens} contains 6{,}000 users and their movie-interaction histories. The task is next-item prediction: given a user profile, past interactions, and candidate items, the policy outputs the predicted movie in JSON. We embed user profiles and apply K-Means to assign cluster ids, and split train/test 2:1 (Appendix~\ref{apdx/movielens}).

\paragraph{Synthetic Preference Dataset.} We define seven personas, each tied to a music-genre preference and a distinct linguistic style (e.g., vocabulary complexity, tone, syntax). Using songs generated per genre by Qwen3-32B, we prompt an LLM to write persona-conditioned reviews for each song. The resulting corpus varies along two axes: sentiment polarity (whether genre matches the persona's preference) and language use (persona-specific style), letting us study their interaction in generated reviews (Appendix~\ref{apdx:SDG}).

\paragraph{Goodreads} \citep{DBLP:conf/acl/WanMNM19} pairs book metadata with user reviews. We curate 10{,}000 reviews (50/50 split), prompting the policy to write a review from the book's title and abstract. As this dataset lacks explicit user profiles, we use the review rating as a proxy cluster id (grouping users by sentiment), and retrieve a length-filtered ground-truth review as reference (Appendix~\ref{apdx/goodreads}).

\paragraph{KGRec} \citep{kgrec-music} contains user interaction sequences over music tracks, each with a text description. The task is to describe the track a user is most likely to play next, given the user profile and descriptions of past interactions. We cluster users via K-Means on embeddings of interacted descriptions and split train/test 50/50 (Appendix~\ref{apdx/kgrec}).

\subsubsection{Models}
We employ two families of LLMs at different sizes as the backbones for our experiments: Gemma-2B, Qwen3-1.7B, and Qwen3-8B. 
We conduct a systematic hyperparameter sweep covering the learning rate, the KL regularization coefficient $\beta$, and the generation group size $G$. To ensure a fair comparison, we maintain consistent hyperparameter configurations across both the baseline GRPO algorithm and our proposed Personalized GRPO algorithm.

\subsubsection{Rewards} For MovieLens-1M, we assign the LLM policy two types of rewards: (1) the correctness of the answer (the model predicting the correct next movie out of several options) and (2) adherence to the JSON format, with relative weights of $1:0.1$. We evaluate top-1 accuracy for testing, consistent with the multiple-choice formulation. For synthetic data, Goodreads, and KGRec, we retrieve textual descriptions of ground-truth items as reference answers and compute rewards based on ROUGE-N and ROUGE-L metrics. We additionally compute cosine similarity between embeddings of generated and reference text, using the \texttt{all-MiniLM-L6-v2} model, to assess semantic similarity.

\subsubsection{Baselines}
We compare against the standard GRPO and GDPO \citep{yao2025no}, an off-policy method that conditions the DPO loss on user clusters. GDPO extends standard preference optimization by introducing cluster-specific loss functions, allowing the model to capture distributional differences across user groups. On MovieLens-1M, we additionally compare against two optimization-level personalization baselines that isolate the two mechanisms of P-GRPO: (1) \emph{Stratified GRPO}, which applies within-batch normalization separately per cluster rather than globally, isolating the benefit of historical (rather than batch-local) statistics; and (2) Variational Preference Learning \citep[VPL;][]{poddar2024personalizing}, which infers latent preference variables via a variational encoder. We further note that our single-cluster ($k{=}1$) configuration is functionally equivalent to global advantage normalization, i.e., REINFORCE++ \citep{hu2025reinforcestabilizingcriticfreepolicy}, as it maintains a single global running mean and variance.

\begin{figure*}[!t]
    \centering
    \includegraphics[width=0.32\linewidth]{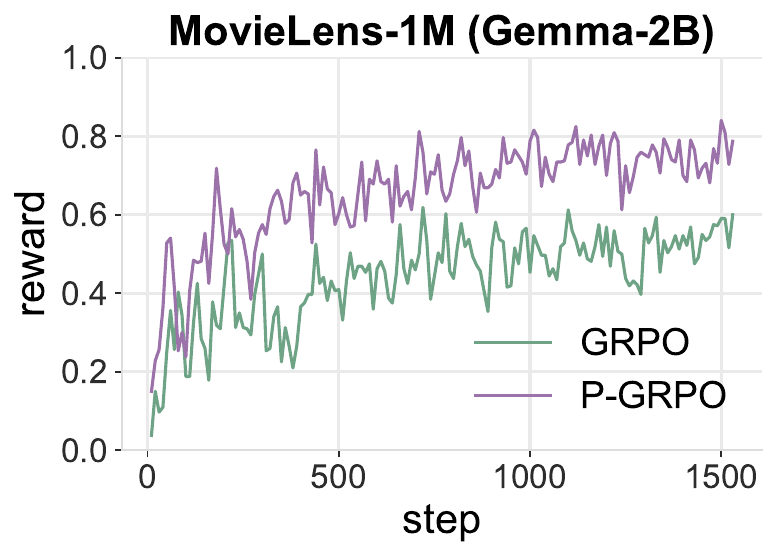}
    \hfill
    \includegraphics[width=0.32\linewidth]{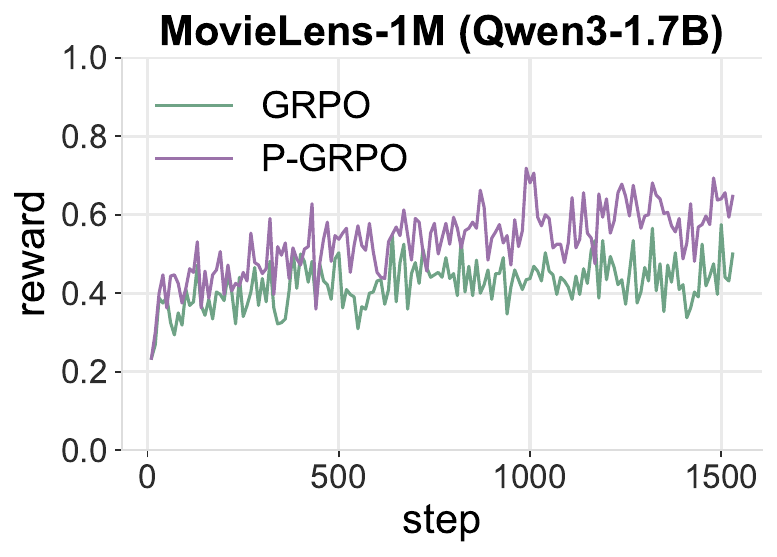}
    \hfill
    \includegraphics[width=0.32\linewidth]{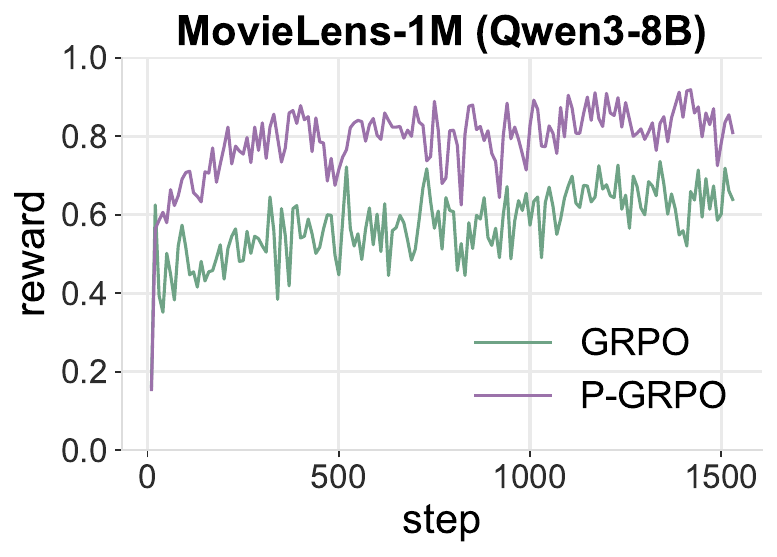}

    \caption{Training reward curves comparing GRPO and P-GRPO on the MovieLens-1M next-item prediction task across three models: Gemma-2B (left), Qwen3-1.7B (center), and Qwen3-8B (right). P-GRPO consistently converges faster and achieves higher average rewards, demonstrating improved learning efficiency through preference-specific normalization.}
    \label{fig:movielens/train/reward}
\end{figure*}

\begin{figure*}
    \centering
    \begin{minipage}[t]{0.32\linewidth}
        \centering
        \includegraphics[width=0.95\linewidth]{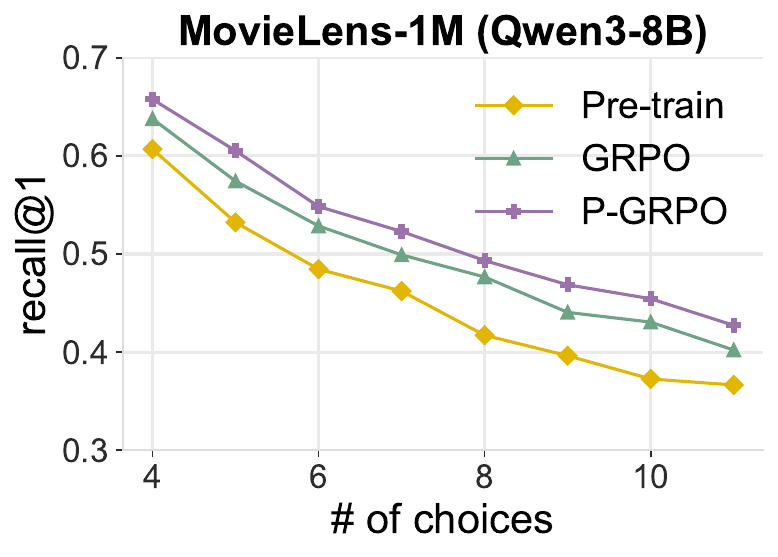}
        \caption{Test accuracy of Qwen3-8B model on MovieLens-1M dataset. 
        P-GRPO consistently outperforms GRPO across all settings.}
        \label{fig:movielens/eval/qwen3}
    \end{minipage}
    \hfill
    \begin{minipage}[t]{0.65\linewidth}
        \centering
        \includegraphics[width=0.48\linewidth]{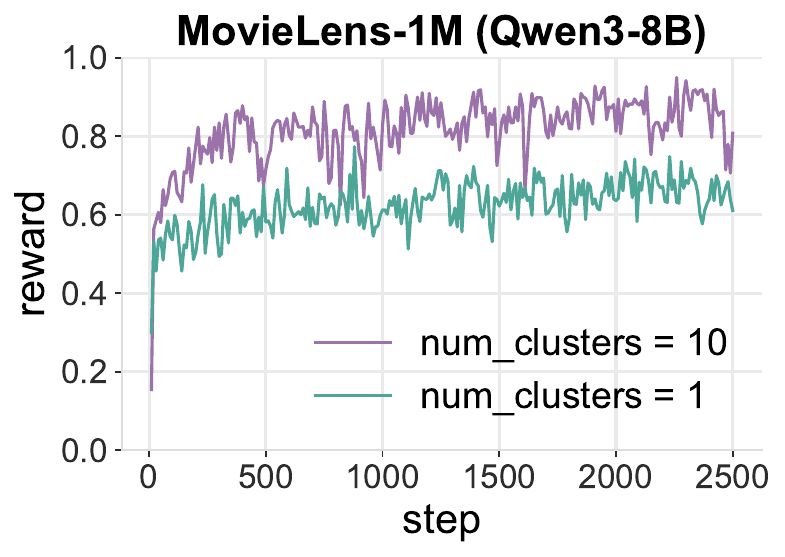}
        \hfill
        \includegraphics[width=0.48\linewidth]{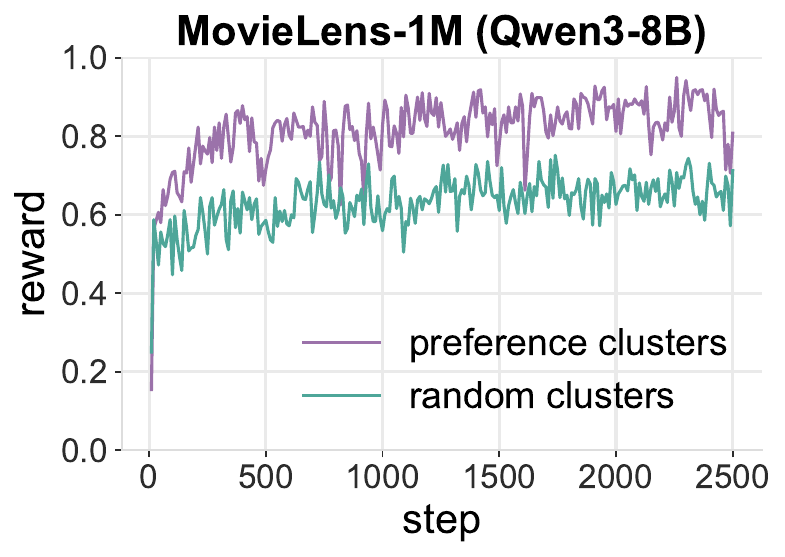}
        \caption{Ablation study on the impact of clustering quality for P-GRPO training on MovieLens-1M dataset with Qwen3-8B. \textbf{Left:}
        Finer-grained clustering achieves higher rewards than coarser clustering. \textbf{Right:} Effect of random cluster assignment versus K-Means clustering, demonstrating that cluster quality is essential for personalization gains.}
        \label{fig:movielens/abation}
    \end{minipage}
\end{figure*}

\subsection{Main Results}

\subsubsection{Content Recommendation}

\paragraph{Training Dynamics.} We present the training reward curves of GRPO and P-GRPO in Figure~\ref{fig:movielens/train/reward}. The comparison demonstrates that: (1) P-GRPO converges faster, attaining stable performance earlier in the training process than GRPO. This behavior indicates that preference-specific normalization yields more stable and informative gradient signals, thereby improving learning efficiency; (2) P-GRPO consistently achieves higher average rewards over the training process, suggesting that explicitly modeling preference-specific biases enables the model to better accommodate diverse user groups instead of disproportionately optimizing for simpler preferences. These results are consistent regardless of model architectures and model sizes.

\paragraph{Testing Performance.}Figure~\ref{fig:movielens/eval/qwen3} compares the evaluated top-1 accuracy of the Qwen3-8B models. During training, only four candidate movies are provided. However, we gradually increase the number of candidates during testing. This design aims to assess the generalization capability of the LLMs, with particular emphasis on behavioral understanding. In addition to the standard GRPO algorithm, we include the pre-trained Qwen3-8B model as a baseline, indicated by the yellow curve. In the simplest setting with four choices, P-GRPO achieves an accuracy of $65.77\%$, outperforming the standard GRPO of $63.79\%$. As the number of choices increases, the test accuracy of P-GRPO declines, but it consistently remains higher than GRPO.

\paragraph{Effects of Cluster Granularity.}
We conduct an ablation study by varying the number of preference clusters. As shown in Figure~\ref{fig:movielens/abation}, we compare the training curves of P-GRPO with different numbers of clusters. The results show that a coarser clustering with only one cluster yields inferior performance relative to the finer-grained configuration with ten clusters. We note that the single-cluster setting does not reduce to vanilla GRPO; it is a global running-mean baseline (equivalent to REINFORCE++), so this gap isolates the contribution of cluster conditioning beyond historical normalization alone. This suggests that insufficient cluster granularity restricts the model's ability to differentiate among user preference groups, whereas a larger number of clusters facilitates more effective personalization. A finer-grained sweep over $k$ (Appendix~\ref{apdx/robustness}) shows accuracy improving with granularity and disparity minimized around $k=20$, which coincides with the optimal $k$ suggested by an independent Silhouette analysis, before sample starvation degrades performance at very large $k$.

\paragraph{Effects of Cluster Quality.}
We fix the number of preference clusters to $k=10$, but randomly assign the cluster IDs. As shown in Figure~\ref{fig:movielens/abation}, this random assignment fails to yield performance gains, even when using the P-GRPO. These results indicate that personalization gains depend critically on the quality of clustering, not merely on the presence of preference clusters. To probe this more finely, we conduct a graded label-noise study, replacing each sample's cluster ID with a random one at rate $\eta \in \{0, 0.25, 0.5, 0.75, 1.0\}$. P-GRPO degrades gracefully as $\eta$ increases and converges to the vanilla GRPO baseline at $\eta=1.0$, exactly as expected since fully random labels reduce P-GRPO to a global running-mean baseline. We report the full results in Appendix~\ref{apdx/robustness}.

\paragraph{Comparison with Personalization Baselines.}
Table~\ref{tab:movielens/baselines} compares P-GRPO against Stratified GRPO and VPL on MovieLens-1M with Qwen3-8B. Beyond top-1 accuracy, we report the cross-cluster disparity $\Delta$, defined as the accuracy gap between the best- and worst-performing clusters, where a lower $\Delta$ indicates more equitable performance across user groups. P-GRPO is the only method whose accuracy gain over vanilla GRPO is statistically significant ($p<0.05$, paired test over test items). While VPL attains competitive aggregate accuracy, its gain is not significant at the $95\%$ level and it incurs a substantially larger disparity ($\Delta=0.232$ vs.\ $0.113$ for P-GRPO), i.e., it improves the average at the cost of widening the gap between groups. Stratified GRPO attains a low disparity but lower accuracy, consistent with the sample-starvation problem of computing per-cluster variance within a single batch when minority clusters are rare. By aggregating over the full training history, P-GRPO yields a stable estimator even for rare clusters, matching the strongest baseline on accuracy while remaining the most equitable.

\begin{table}[!t]
    \centering
    \caption{Comparison with personalization baselines on MovieLens-1M (Qwen3-8B). We report top-1 accuracy, the $p$-value of a significance test against vanilla GRPO, and the cross-cluster disparity $\Delta$ (accuracy gap between the best- and worst-performing clusters; lower is more equitable).}
    \label{tab:movielens/baselines}
    \small
    \begin{tabular}{@{}lccc@{}}
        \toprule
        \textbf{Method} & \textbf{Accuracy} & \textbf{$p$-value} & \textbf{Cluster $\Delta$} \\
        \midrule
        GRPO            & 0.7190 & --    & 0.1247 \\
        Stratified GRPO & 0.7366 & 0.085 & 0.1120 \\
        VPL             & 0.7531 & 0.054 & 0.2320 \\
        P-GRPO          & $\mathbf{0.7566}$ & $\mathbf{0.009}$ & 0.1130 \\
        \bottomrule
    \end{tabular}
\end{table}

\begin{table*}[!t]
    \centering
    \caption{Test-set generation performance on synthetic, KGRec, and Goodreads datasets. We compare GDPO, GRPO, and P-GRPO using Qwen3-8B and Gemma-2B models. Metrics include ROUGE scores and cosine similarity (Cos. Sim.). Bold indicates best performance per model-dataset combination.}
    \label{tab:result/generation/performance}
    \small
    \resizebox{\textwidth}{!}{
        \begin{tabular}{@{}ll cccc cccc@{}}
            \toprule
            & & \multicolumn{4}{c}{\textbf{Qwen3-8B}} & \multicolumn{4}{c}{\textbf{Gemma-2B}} \\
            \cmidrule(lr){3-6} \cmidrule(lr){7-10}
            \textbf{Dataset} & \textbf{Method} & ROUGE-1 & ROUGE-2 & ROUGE-L & Cos. Sim & ROUGE-1 & ROUGE-2 & ROUGE-L & Cos. Sim \\
            \midrule
            \multirow{3}{*}{Synthetic Data}
            & GDPO & 0.5663 & 0.1374 & 0.3359 & 0.5246 & 0.5297 & 0.1329 & 0.2907 & 0.7050 \\
            & GRPO & 0.6075 & 0.1713 & 0.3624 & 0.5814 & 0.6042 & 0.1775 & 0.3706 & 0.7090 \\
            & P-GRPO & $\mathbf{0.6267}$ & $\mathbf{0.1853}$ & $\mathbf{0.3758}$ & $\mathbf{0.6150}$ & $\mathbf{0.6133}$ & $\mathbf{0.1861}$ & $\mathbf{0.3798}$ & $\mathbf{0.7154}$ \\
            \midrule
            \multirow{3}{*}{Goodreads}
            & GDPO & 0.4987 & 0.0781 & $0.3003$ & 0.4520 & 0.4008 & 0.0600 &  0.2254 & 0.5288 \\
            & GRPO & 0.6076 & 0.1374 & $\mathbf{0.3596}$ & 0.5266 & $0.5526$ & 0.1060 & $\mathbf{0.3255}$ & $0.5328$\\
            & P-GRPO & $\mathbf{0.6138}$ & $\mathbf{0.1383}$ & $0.3587$ & $\mathbf{0.5296}$ & $\mathbf{0.5534}$ & $\mathbf{0.1181}$ & $0.3204$ & $\mathbf{0.5415}$ \\
            \midrule
            \multirow{3}{*}{KGRec}
            & GDPO & 0.3845 & 0.0473 & 0.2419 & $\mathbf{0.3177}$ & 0.2513 & 0.0226 & $0.1455$ & 0.3088 \\
            & GRPO & 0.4340 & 0.0790 & $\mathbf{0.2848}$ & 0.2905 & 0.5603 & 0.1058 & $0.2832$ & 0.3069 \\
            & P-GRPO & $\mathbf{0.4649}$ & $\mathbf{0.1067}$ & $0.2840$ & 0.2907 & $\mathbf{0.5618}$ & $\mathbf{0.1067}$ & $\mathbf{0.2843}$ & $\mathbf{0.3130}$ \\
            \bottomrule
        \end{tabular}
    }
\end{table*}

\subsubsection{Language Generation}
Table~\ref{tab:result/generation/performance} presents a breakdown of the test-set performance comparing GDPO, GRPO and P-GRPO across the three preference-aligned language generation datasets (see Appendix~\ref{apdx/add_exp} for detailed results including standard deviations).
\begin{itemize}[leftmargin=*]
\item \textbf{Synthetic Preference Dataset.} For the synthetic data experiments, P-GRPO demonstrates improvements across all metrics for both model architectures. For Qwen3-8B, P-GRPO achieves a ROUGE-1 score of $0.6267$ compared to $0.6075$ for GRPO, while ROUGE-2 increases from $0.1713$ to $0.1853$. Notably, the cosine similarity metric shows an improvement from $0.5814$ to $0.6150$, indicating better semantic alignment with ground-truth outputs. We also observed that Gemma-2B exhibits similar patterns, with ROUGE-1 improving from $0.6042$ to $0.6133$ and ROUGE-2 from $0.1775$ to $0.1861$.

\item \textbf{Goodreads.} For Goodreads, the proposed P-GRPO algorithm yields improvements for the Qwen3-8B model. We observe a consistent upward trend for ROUGE-1 and ROUGE-2 scores. For the smaller Gemma-2B model, the ROUGE-2 score improves from $0.1060$ to $0.1181$, though the ROUGE-1 improvement is more modest ($0.5526$ to $0.5534$). We observe that for both models the ROUGE-L scores show slight decreases. However, the overall quality improves as evidenced by higher cosine similarity. 
We also identify that the GDPO algorithm failed to generate reviews that aligned with the references. This result again suggests that off-policy reinforcement learning performs poorly on this task.

\item \textbf{KGRec.} For KGRec, with the Qwen3-8B model, the ROUGE-1 score improves from $0.4340$ to $0.4649$, while ROUGE-2 improves from $0.0790$ to $0.1067$, reflecting significantly stronger n-gram overlap with the reference texts.  With Gemma-2B, P-GRPO maintains comparable ROUGE-1 performance ($0.5618$ vs. $0.5603$) while achieving the highest ROUGE-2 ($0.1067$) and cosine similarity ($0.3130$) scores among all methods.
\end{itemize}

\begin{figure}[!t]
    \centering
    \includegraphics[width=0.8\linewidth]{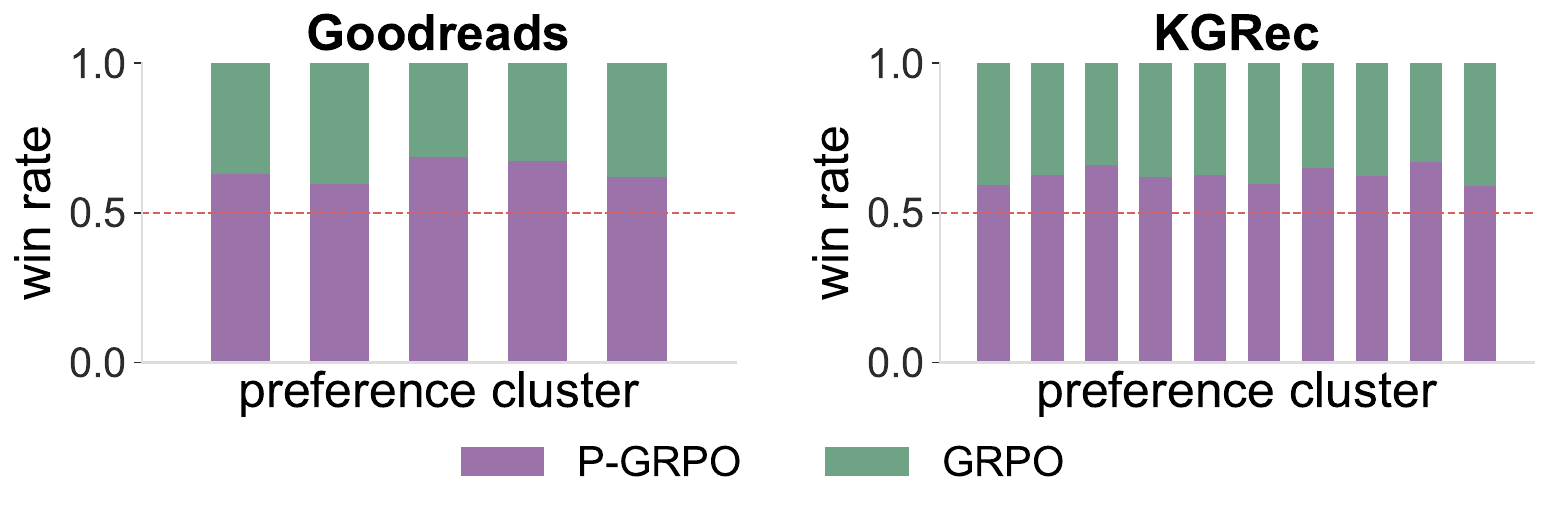}
    \caption{LLM-as-judge win rates across user preference clusters. Using GPT-OSS-120B as the judge, we compare responses generated by P-GRPO versus GRPO based on semantic quality, coherence, and user preference alignment. P-GRPO achieves higher win rates across all clusters in both datasets, demonstrating superior personalized generation capabilities.}
    \label{fig:result/win_rates}
\end{figure}

To further analyze generation quality, we conduct an LLM-as-judge evaluation on the KGRec and Goodreads datasets, using the GPT-OSS-120B \citep{openai2025gptoss120bgptoss20bmodel} as the judge. For each instance, we present the judge with the original input prompt, the user profile context, the reference answer, and two candidate texts from the GRPO- and P-GRPO-trained models, without revealing which method produced each output. The judge scores the pair along semantic quality, coherence, and user preference alignment. Aggregating win rates across the preference clusters used during training (Figure~\ref{fig:result/win_rates}), P-GRPO beats GRPO in every cluster on both datasets, confirming a stronger ability to generate personalized output aligned with individual user preferences. We also run a second independent judge to test the robustness of judge, which corroborates the same ranking (Appendix~\ref{apdx/judge-robustness}).

\section{Conclusion}

In this work, we addressed a fundamental limitation in current LLM alignment methods: the inability to effectively accommodate heterogeneous user preferences. 
To address this limitation, we proposed Personalized GRPO (P-GRPO), which reconceptualizes advantage normalization by maintaining preference-specific historical statistics rather than normalizing within generation groups. We validated P-GRPO across three distinct personalization environments spanning behavior modeling and review generation, across different model architectures and sizes. Our experiments consistently demonstrated that P-GRPO achieves faster convergence and outperforms the standard GRPO approach. Importantly, our approach achieves these improvements purely stemming from more principled advantage normalization that accounts for preference heterogeneity.

While P-GRPO demonstrates promising results, several limitations warrant further investigation. Firstly, our approach assumes that preference groups can be meaningfully partitioned and remain unchanged over time. In practice, user preferences may evolve dynamically, requiring test-time clustering strategies or continual learning mechanisms. Second, our experiments primarily focused on recommendation and generation tasks; exploring P-GRPO's effectiveness in other domains would provide valuable insights into its generalization capability.

\section*{Ethics Statement}
P-GRPO is motivated in part by an equity concern in current LLM alignment, namely the systematic suppression of minority preference signals. However, personalization itself introduces ethical risks that warrant careful consideration.

While personalization can improve user experience, it also risks reducing exposure to different perspectives and contributing to societal polarization. Users may receive responses that increasingly reflect their existing viewpoints. In domains involving opinion or values, excessive personalization could fragment shared informational commons, and bad actors could exploit personalization mechanisms to deliver targeted misinformation tailored to individual vulnerabilities.

The quality of personalization depends critically on the clustering mechanism, raising additional fairness concerns. Clustering algorithms may create or reinforce problematic categorizations that do not reflect users' actual preferences or identities, and users at cluster boundaries may receive suboptimal personalization. The current approach assumes preference groups remain stable over time, but failure to accommodate preference drift could disadvantage users whose needs change. Furthermore, while P-GRPO aims to provide equitable optimization across preference groups, uneven data availability could still result in quality disparities, and standard benchmarks may not adequately capture performance differences across groups.

We recommend several measures for responsible deployment. Future implementations should explore privacy-preserving approaches such as federated learning or differential privacy that enable personalization without centralized storage of preference data. Users should have visibility into their assigned preference group, the ability to modify their classification, and clear options to opt out of personalization. Practitioners should conduct disaggregated evaluations across preference groups to identify and mitigate performance disparities. Extensions of P-GRPO should incorporate mechanisms for detecting and adapting to preference drift. Finally, careful consideration should be given to which domains benefit from personalization versus those where consistent model behavior is preferable, such as factual information or safety-critical applications.


\bibliography{include/example_paper}

@inproceedings{DBLP:conf/acl/WanMNM19,
  author       = {Mengting Wan and
                  Rishabh Misra and
                  Ndapa Nakashole and
                  Julian J. McAuley},
  editor       = {Anna Korhonen and
                  David R. Traum and
                  Llu{\'{\i}}s M{\`{a}}rquez},
  title        = {Fine-Grained Spoiler Detection from Large-Scale Review Corpora},
  booktitle    = {Proceedings of the 57th Conference of the Association for Computational
                  Linguistics, {ACL} 2019, Florence, Italy, July 28- August 2, 2019,
                  Volume 1: Long Papers},
  pages        = {2605--2610},
  publisher    = {Association for Computational Linguistics},
  year         = {2019},
  doi          = {10.18653/V1/P19-1248},
  bibsource    = {dblp computer science bibliography, https://dblp.org}
}

@article{kgrec-music,
author = {Oramas, Sergio and Ostuni, Vito Claudio and Noia, Tommaso Di and Serra, Xavier and Sciascio, Eugenio Di},
title = {Sound and Music Recommendation with Knowledge Graphs},
year = {2016},
issue_date = {March 2017},
publisher = {Association for Computing Machinery},
address = {New York, NY, USA},
volume = {8},
number = {2},
issn = {2157-6904},
url = {https://doi.org/10.1145/2926718},
doi = {10.1145/2926718},
journal = {ACM Trans. Intell. Syst. Technol.},
month = oct,
articleno = {21},
numpages = {21}
}

@article{konstan1997grouplens,
  author = {Konstan, Joseph A. and Resnick, Paul and Riedl, John and Borchers, Axel and Barry, John},
  title = {{GroupLens}: composing collaboration from fine-grained reactions to arguments},
  journal = {Commun. ACM},
  issue_date = {Dec. 1997},
  volume = {40},
  number = {12},
  pages = {32--43},
  articleno = {F500001--4},
  numpages = {12},
  year = {1997},
  publisher = {Association for Computing Machinery},
  address = {New York, NY, USA},
  issn = {0001-0782},
  url = {doi.org}
}

@article{
zhang2025personalization,
title={Personalization of Large Language Models: A Survey},
author={Zhehao Zhang and Ryan A. Rossi and Branislav Kveton and Yijia Shao and Diyi Yang and Hamed Zamani and Franck Dernoncourt and Joe Barrow and Tong Yu and Sungchul Kim and Ruiyi Zhang and Jiuxiang Gu and Tyler Derr and Hongjie Chen and Junda Wu and Xiang Chen and Zichao Wang and Subrata Mitra and Nedim Lipka and Nesreen K. Ahmed and Yu Wang},
journal={Transactions on Machine Learning Research},
issn={2835-8856},
year={2025},
url={https://openreview.net/forum?id=tf6A9EYMo6},
note={Survey Certification}
}

@inproceedings{guan-etal-2025-survey,
    title = "A Survey on Personalized {A}lignment{---}{T}he Missing Piece for Large Language Models in Real-World Applications",
    author = "Guan, Jian  and
      Wu, Junfei  and
      Li, Jia-Nan  and
      Cheng, Chuanqi  and
      Wu, Wei",
    editor = "Che, Wanxiang  and
      Nabende, Joyce  and
      Shutova, Ekaterina  and
      Pilehvar, Mohammad Taher",
    booktitle = "Findings of the Association for Computational Linguistics: ACL 2025",
    month = jul,
    year = "2025",
    address = "Vienna, Austria",
    publisher = "Association for Computational Linguistics",
    url = "https://aclanthology.org/2025.findings-acl.277/",
    doi = "10.18653/v1/2025.findings-acl.277",
    pages = "5313--5333",
    ISBN = "979-8-89176-256-5"
}

@article{jang2023personalized,
  title={Personalized Soups: Personalized Large Language Model Alignment via Post-hoc Parameter Merging},
  author={Jang, Joel and Kim, Seungone and Lin, Bill Yuchen and Wang, Yizhong and Hessel, Jack and Zettlemoyer, Luke and Hajishirzi, Hannaneh and Choi, Yejin and Ammanabrolu, Prithviraj},
  journal={arXiv preprint arXiv:2310.11564},
  year={2023}
}

@inproceedings{
poddar2024personalizing,
title={Personalizing Reinforcement Learning from Human Feedback with Variational Preference Learning},
author={Sriyash Poddar and Yanming Wan and Hamish Ivison and Abhishek Gupta and Natasha Jaques},
booktitle={The Thirty-eighth Annual Conference on Neural Information Processing Systems},
year={2024},
url={https://openreview.net/forum?id=gRG6SzbW9p}
}

@article{personalizedRLHF,
  title={Personalized language modeling from personalized human feedback},
  author={Li, Xinyu and Zhou, Ruiyang and Lipton, Zachary C and Leqi, Liu},
  journal={arXiv preprint arXiv:2402.05133},
  year={2024}
}

@inproceedings{
yao2025no,
title={No Preference Left Behind: Group Distributional Preference Optimization},
author={Binwei Yao and Zefan Cai and Yun-Shiuan Chuang and Shanglin Yang and Ming Jiang and Diyi Yang and Junjie Hu},
booktitle={The Thirteenth International Conference on Learning Representations},
year={2025},
url={https://openreview.net/forum?id=bgpNJBD6Va}
}

@misc{singh2025fspofewshotpreferenceoptimization,
      title={FSPO: Few-Shot Preference Optimization of Synthetic Preference Data in LLMs Elicits Effective Personalization to Real Users}, 
      author={Anikait Singh and Sheryl Hsu and Kyle Hsu and Eric Mitchell and Stefano Ermon and Tatsunori Hashimoto and Archit Sharma and Chelsea Finn},
      year={2025},
      eprint={2502.19312},
      archivePrefix={arXiv},
      primaryClass={cs.LG},
      url={https://arxiv.org/abs/2502.19312}, 
}

@misc{dipalma2024evaluatingchatgptrecommendersystem,
      title={Evaluating ChatGPT as a Recommender System: A Rigorous Approach}, 
      author={Dario Di Palma and Giovanni Maria Biancofiore and Vito Walter Anelli and Fedelucio Narducci and Tommaso Di Noia and Eugenio Di Sciascio},
      year={2024},
      eprint={2309.03613},
      archivePrefix={arXiv},
      primaryClass={cs.IR},
      url={https://arxiv.org/abs/2309.03613}, 
}

@inproceedings{
jiang2023evaluating,
title={Evaluating and Inducing Personality in Pre-trained Language Models},
author={Guangyuan Jiang and Manjie Xu and Song-Chun Zhu and Wenjuan Han and Chi Zhang and Yixin Zhu},
booktitle={Thirty-seventh Conference on Neural Information Processing Systems},
year={2023},
url={https://openreview.net/forum?id=I9xE1Jsjfx}
}

@misc{richardson2023integratingsummarizationretrievalenhanced,
      title={Integrating Summarization and Retrieval for Enhanced Personalization via Large Language Models}, 
      author={Chris Richardson and Yao Zhang and Kellen Gillespie and Sudipta Kar and Arshdeep Singh and Zeynab Raeesy and Omar Zia Khan and Abhinav Sethy},
      year={2023},
      eprint={2310.20081},
      archivePrefix={arXiv},
      primaryClass={cs.CL},
      url={https://arxiv.org/abs/2310.20081}, 
}

@inproceedings{10.1145/3701716.3715463,
author = {Ning, Lin and Liu, Luyang and Wu, Jiaxing and Wu, Neo and Berlowitz, Devora and Prakash, Sushant and Green, Bradley and O'Banion, Shawn and Xie, Jun},
title = {User-LLM: Efficient LLM Contextualization with User Embeddings},
year = {2025},
isbn = {9798400713316},
publisher = {Association for Computing Machinery},
address = {New York, NY, USA},
url = {https://doi.org/10.1145/3701716.3715463},
doi = {10.1145/3701716.3715463},
booktitle = {Companion Proceedings of the ACM on Web Conference 2025},
pages = {1219–1223},
numpages = {5},
location = {Sydney NSW, Australia},
series = {WWW '25}
}

@book{johnsen2024large,
  title={Large language models (LLMs)},
  author={Johnsen, Maria},
  year={2024},
  publisher={Maria Johnsen}
}

@article{ouyang2022training,
  title={Training language models to follow instructions with human feedback},
  author={Ouyang, Long and Wu, Jeffrey and Jiang, Xu and Almeida, Diogo and Wainwright, Carroll L and Mishkin, Pamela and Zhang, Chong and Agarwal, Sandhini and Slama, Katarina and Ray, Alex and others},
  journal={Advances in Neural Information Processing Systems},
  volume={35},
  pages={27730--27744},
  year={2022}
}

@misc{wang2024reinforcementlearningenhancedllms,
      title={Reinforcement Learning Enhanced LLMs: A Survey}, 
      author={Shuhe Wang and Shengyu Zhang and Jie Zhang and Runyi Hu and Xiaoya Li and Tianwei Zhang and Jiwei Li and Fei Wu and Guoyin Wang and Eduard Hovy},
      year={2024},
      eprint={2412.10400},
      archivePrefix={arXiv},
      primaryClass={cs.CL},
      url={https://arxiv.org/abs/2412.10400}, 
}

@misc{deepseekai2025deepseekr1incentivizingreasoningcapability,
      title={DeepSeek-R1: Incentivizing Reasoning Capability in LLMs via Reinforcement Learning}, 
      author={DeepSeek-AI},
      year={2025},
      eprint={2501.12948},
      archivePrefix={arXiv},
      primaryClass={cs.CL},
      url={https://arxiv.org/abs/2501.12948}, 
}

@misc{deepseek-math,
  publtype={informal},
  author={Zhihong Shao and Peiyi Wang and Qihao Zhu and Runxin Xu and Junxiao Song and Mingchuan Zhang and Y. K. Li and Y. Wu and Daya Guo},
  title={DeepSeekMath: Pushing the Limits of Mathematical Reasoning in Open Language Models},
  year={2024},
  cdate={1704067200000},
  journal={CoRR},
  volume={abs/2402.03300},
  url={https://doi.org/10.48550/arXiv.2402.03300}
}

@inproceedings{
yu2025dapo,
title={{DAPO}: An Open-Source {LLM} Reinforcement Learning System at Scale},
author={Qiying Yu and Zheng Zhang and Ruofei Zhu and Yufeng Yuan and Xiaochen Zuo and YuYue and Weinan Dai and Tiantian Fan and Gaohong Liu and Juncai Liu and LingJun Liu and Xin Liu and Haibin Lin and Zhiqi Lin and Bole Ma and Guangming Sheng and Yuxuan Tong and Chi Zhang and Mofan Zhang and Ru Zhang and Wang Zhang and Hang Zhu and Jinhua Zhu and Jiaze Chen and Jiangjie Chen and Chengyi Wang and Hongli Yu and Yuxuan Song and Xiangpeng Wei and Hao Zhou and Jingjing Liu and Wei-Ying Ma and Ya-Qin Zhang and Lin Yan and Yonghui Wu and Mingxuan Wang},
booktitle={The Thirty-ninth Annual Conference on Neural Information Processing Systems},
year={2025},
url={https://openreview.net/forum?id=2a36EMSSTp}
}

@article{holtgraves1997styles,
  title={Styles of language use: Individual and cultural variability in conversational indirectness},
  author={Holtgraves, Thomas},
  journal={Journal of Personality and Social Psychology},
  volume={73},
  number={3},
  pages={624--637},
  year={1997},
  publisher={American Psychological Association},
  doi={10.1037/0022-3514.73.3.624}
}

@article{mccrae1987validation,
  title={Validation of the five-factor model of personality across instruments and observers},
  author={McCrae, Robert R and Costa Jr, Paul T},
  journal={Journal of Personality and Social Psychology},
  volume={52},
  number={1},
  pages={81--90},
  year={1987},
  publisher={American Psychological Association},
  doi={10.1037/0022-3514.52.1.81}
}

@article{qwen3,
    title={Qwen3 Technical Report}, 
    author={An Yang and Anfeng Li and Baosong Yang and Beichen Zhang and Binyuan Hui and Bo Zheng and Bowen Yu and Chang Gao and Chengen Huang and Chenxu Lv and Chujie Zheng and Dayiheng Liu and Fan Zhou and Fei Huang and Feng Hu and Hao Ge and Haoran Wei and Huan Lin and Jialong Tang and Jian Yang and Jianhong Tu and Jianwei Zhang and Jianxin Yang and Jiaxi Yang and Jing Zhou and Jingren Zhou and Junyang Lin and Kai Dang and Keqin Bao and Kexin Yang and Le Yu and Lianghao Deng and Mei Li and Mingfeng Xue and Mingze Li and Pei Zhang and Peng Wang and Qin Zhu and Rui Men and Ruize Gao and Shixuan Liu and Shuang Luo and Tianhao Li and Tianyi Tang and Wenbiao Yin and Xingzhang Ren and Xinyu Wang and Xinyu Zhang and Xuancheng Ren and Yang Fan and Yang Su and Yichang Zhang and Yinger Zhang and Yu Wan and Yuqiong Liu and Zekun Wang and Zeyu Cui and Zhenru Zhang and Zhipeng Zhou and Zihan Qiu},
    journal = {arXiv preprint arXiv:2505.09388},
    year={2025}
}

@misc{gemmateam2024gemma2improvingopen,
      title={Gemma 2: Improving Open Language Models at a Practical Size}, 
      author={Gemma Team and Morgane Riviere and Shreya Pathak and Pier Giuseppe Sessa and Cassidy Hardin and Surya Bhupatiraju and Léonard Hussenot and Thomas Mesnard and Bobak Shahriari and Alexandre Ramé and Johan Ferret and Peter Liu and Pouya Tafti and Abe Friesen and Michelle Casbon and Sabela Ramos and Ravin Kumar and Charline Le Lan and Sammy Jerome and Anton Tsitsulin and Nino Vieillard and Piotr Stanczyk and Sertan Girgin and Nikola Momchev and Matt Hoffman and Shantanu Thakoor and Jean-Bastien Grill and Behnam Neyshabur and Olivier Bachem and Alanna Walton and Aliaksei Severyn and Alicia Parrish and Aliya Ahmad and Allen Hutchison and Alvin Abdagic and Amanda Carl and Amy Shen and Andy Brock and Andy Coenen and Anthony Laforge and Antonia Paterson and Ben Bastian and Bilal Piot and Bo Wu and Brandon Royal and Charlie Chen and Chintu Kumar and Chris Perry and Chris Welty and Christopher A. Choquette-Choo and Danila Sinopalnikov and David Weinberger and Dimple Vijaykumar and Dominika Rogozińska and Dustin Herbison and Elisa Bandy and Emma Wang and Eric Noland and Erica Moreira and Evan Senter and Evgenii Eltyshev and Francesco Visin and Gabriel Rasskin and Gary Wei and Glenn Cameron and Gus Martins and Hadi Hashemi and Hanna Klimczak-Plucińska and Harleen Batra and Harsh Dhand and Ivan Nardini and Jacinda Mein and Jack Zhou and James Svensson and Jeff Stanway and Jetha Chan and Jin Peng Zhou and Joana Carrasqueira and Joana Iljazi and Jocelyn Becker and Joe Fernandez and Joost van Amersfoort and Josh Gordon and Josh Lipschultz and Josh Newlan and Ju-yeong Ji and Kareem Mohamed and Kartikeya Badola and Kat Black and Katie Millican and Keelin McDonell and Kelvin Nguyen and Kiranbir Sodhia and Kish Greene and Lars Lowe Sjoesund and Lauren Usui and Laurent Sifre and Lena Heuermann and Leticia Lago and Lilly McNealus and Livio Baldini Soares and Logan Kilpatrick and Lucas Dixon and Luciano Martins and Machel Reid and Manvinder Singh and Mark Iverson and Martin Görner and Mat Velloso and Mateo Wirth and Matt Davidow and Matt Miller and Matthew Rahtz and Matthew Watson and Meg Risdal and Mehran Kazemi and Michael Moynihan and Ming Zhang and Minsuk Kahng and Minwoo Park and Mofi Rahman and Mohit Khatwani and Natalie Dao and Nenshad Bardoliwalla and Nesh Devanathan and Neta Dumai and Nilay Chauhan and Oscar Wahltinez and Pankil Botarda and Parker Barnes and Paul Barham and Paul Michel and Pengchong Jin and Petko Georgiev and Phil Culliton and Pradeep Kuppala and Ramona Comanescu and Ramona Merhej and Reena Jana and Reza Ardeshir Rokni and Rishabh Agarwal and Ryan Mullins and Samaneh Saadat and Sara Mc Carthy and Sarah Cogan and Sarah Perrin and Sébastien M. R. Arnold and Sebastian Krause and Shengyang Dai and Shruti Garg and Shruti Sheth and Sue Ronstrom and Susan Chan and Timothy Jordan and Ting Yu and Tom Eccles and Tom Hennigan and Tomas Kocisky and Tulsee Doshi and Vihan Jain and Vikas Yadav and Vilobh Meshram and Vishal Dharmadhikari and Warren Barkley and Wei Wei and Wenming Ye and Woohyun Han and Woosuk Kwon and Xiang Xu and Zhe Shen and Zhitao Gong and Zichuan Wei and Victor Cotruta and Phoebe Kirk and Anand Rao and Minh Giang and Ludovic Peran and Tris Warkentin and Eli Collins and Joelle Barral and Zoubin Ghahramani and Raia Hadsell and D. Sculley and Jeanine Banks and Anca Dragan and Slav Petrov and Oriol Vinyals and Jeff Dean and Demis Hassabis and Koray Kavukcuoglu and Clement Farabet and Elena Buchatskaya and Sebastian Borgeaud and Noah Fiedel and Armand Joulin and Kathleen Kenealy and Robert Dadashi and Alek Andreev},
      year={2024},
      eprint={2408.00118},
      archivePrefix={arXiv},
      primaryClass={cs.CL},
      url={https://arxiv.org/abs/2408.00118}, 
}

@article{gspo,
  title={Group Sequence Policy Optimization}, 
  author={
    Chujie Zheng and Shixuan Liu and Mingze Li and Xiong-Hui Chen and Bowen Yu and 
    Chang Gao and Kai Dang and Yuqiong Liu and Rui Men and An Yang and Jingren Zhou and 
    Junyang Lin 
  },
  journal={arXiv preprint arXiv:2507.18071},
  year={2025}
}

@misc{openai2025gptoss120bgptoss20bmodel,
      title={gpt-oss-120b \& gpt-oss-20b Model Card}, 
      author={OpenAI and : and Sandhini Agarwal and Lama Ahmad and Jason Ai and Sam Altman and Andy Applebaum and Edwin Arbus and Rahul K. Arora and Yu Bai and Bowen Baker and Haiming Bao and Boaz Barak and Ally Bennett and Tyler Bertao and Nivedita Brett and Eugene Brevdo and Greg Brockman and Sebastien Bubeck and Che Chang and Kai Chen and Mark Chen and Enoch Cheung and Aidan Clark and Dan Cook and Marat Dukhan and Casey Dvorak and Kevin Fives and Vlad Fomenko and Timur Garipov and Kristian Georgiev and Mia Glaese and Tarun Gogineni and Adam Goucher and Lukas Gross and Katia Gil Guzman and John Hallman and Jackie Hehir and Johannes Heidecke and Alec Helyar and Haitang Hu and Romain Huet and Jacob Huh and Saachi Jain and Zach Johnson and Chris Koch and Irina Kofman and Dominik Kundel and Jason Kwon and Volodymyr Kyrylov and Elaine Ya Le and Guillaume Leclerc and James Park Lennon and Scott Lessans and Mario Lezcano-Casado and Yuanzhi Li and Zhuohan Li and Ji Lin and Jordan Liss and Lily and Liu and Jiancheng Liu and Kevin Lu and Chris Lu and Zoran Martinovic and Lindsay McCallum and Josh McGrath and Scott McKinney and Aidan McLaughlin and Song Mei and Steve Mostovoy and Tong Mu and Gideon Myles and Alexander Neitz and Alex Nichol and Jakub Pachocki and Alex Paino and Dana Palmie and Ashley Pantuliano and Giambattista Parascandolo and Jongsoo Park and Leher Pathak and Carolina Paz and Ludovic Peran and Dmitry Pimenov and Michelle Pokrass and Elizabeth Proehl and Huida Qiu and Gaby Raila and Filippo Raso and Hongyu Ren and Kimmy Richardson and David Robinson and Bob Rotsted and Hadi Salman and Suvansh Sanjeev and Max Schwarzer and D. Sculley and Harshit Sikchi and Kendal Simon and Karan Singhal and Yang Song and Dane Stuckey and Zhiqing Sun and Philippe Tillet and Sam Toizer and Foivos Tsimpourlas and Nikhil Vyas and Eric Wallace and Xin Wang and Miles Wang and Olivia Watkins and Kevin Weil and Amy Wendling and Kevin Whinnery and Cedric Whitney and Hannah Wong and Lin Yang and Yu Yang and Michihiro Yasunaga and Kristen Ying and Wojciech Zaremba and Wenting Zhan and Cyril Zhang and Brian Zhang and Eddie Zhang and Shengjia Zhao},
      year={2025},
      eprint={2508.10925},
      archivePrefix={arXiv},
      primaryClass={cs.CL},
      url={https://arxiv.org/abs/2508.10925}, 
}

@inproceedings{
pan2022the,
title={The Effects of Reward Misspecification: Mapping and Mitigating Misaligned Models},
author={Alexander Pan and Kush Bhatia and Jacob Steinhardt},
booktitle={International Conference on Learning Representations},
year={2022},
url={https://openreview.net/forum?id=JYtwGwIL7ye}
}

@misc{macdiarmid2025naturalemergentmisalignmentreward,
      title={Natural Emergent Misalignment from Reward Hacking in Production RL}, 
      author={Monte MacDiarmid and Benjamin Wright and Jonathan Uesato and Joe Benton and Jon Kutasov and Sara Price and Naia Bouscal and Sam Bowman and Trenton Bricken and Alex Cloud and Carson Denison and Johannes Gasteiger and Ryan Greenblatt and Jan Leike and Jack Lindsey and Vlad Mikulik and Ethan Perez and Alex Rodrigues and Drake Thomas and Albert Webson and Daniel Ziegler and Evan Hubinger},
      year={2025},
      eprint={2511.18397},
      archivePrefix={arXiv},
      primaryClass={cs.AI},
      url={https://arxiv.org/abs/2511.18397}, 
}

@article{Welford1962,
  author = {B. P. Welford},
  title = {Note on a Method for Calculating Corrected Sums of Squares and Products},
  journal = {Technometrics},
  volume = {4},
  number = {3},
  pages = {419-420},
  year = {1962},
  doi = {10.1080/00401706.1962.10490022}
}

@inproceedings{
zhang2025diverging,
title={Diverging Preferences: When do Annotators Disagree and do Models Know?},
author={Michael JQ Zhang and Zhilin Wang and Jena D. Hwang and Yi Dong and Olivier Delalleau and Yejin Choi and Eunsol Choi and Xiang Ren and Valentina Pyatkin},
booktitle={Forty-second International Conference on Machine Learning},
year={2025},
url={https://openreview.net/forum?id=qWgAAVhoXb}
}

@inproceedings{10.1145/3757887.3767678,
author = {Shirali, Ali and Nasr-Esfahany, Arash and Alomar, Abdullah Omar and Mirtaheri, Parsa and Abebe, Rediet and Procaccia, Ariel D.},
title = {Direct Alignment with Heterogeneous Preferences},
year = {2025},
isbn = {9798400721403},
publisher = {Association for Computing Machinery},
address = {New York, NY, USA},
url = {https://doi.org/10.1145/3757887.3767678},
doi = {10.1145/3757887.3767678},
booktitle = {Proceedings of the 5th ACM Conference on Equity and Access in Algorithms, Mechanisms, and Optimization},
pages = {292},
numpages = {1},
location = {
},
series = {EAAMO '25}
}

@misc{zafari2018modellinganalysistemporalpreference,
      title={Modelling and Analysis of Temporal Preference Drifts Using A Component-Based Factorised Latent Approach}, 
      author={F. Zafari and I. Moser and T. Baarslag},
      year={2018},
      eprint={1802.09728},
      archivePrefix={arXiv},
      primaryClass={cs.IR},
      url={https://arxiv.org/abs/1802.09728}, 
}

@misc{
son2025right,
title={Right Now, Wrong Then: Non-Stationary Direct Preference Optimization under Preference Drift},
author={Seongho Son and William Bankes and Sayak Ray Chowdhury and Brooks Paige and Ilija Bogunovic},
year={2025},
url={https://openreview.net/forum?id=PabAln0jjB}
}

@article{10.1093/pnasnexus/pgae231,
    author = {Peters, Heinrich and Matz, Sandra C},
    title = {Large language models can infer psychological dispositions of social media users},
    journal = {PNAS Nexus},
    volume = {3},
    number = {6},
    pages = {pgae231},
    year = {2024},
    month = {06},
    issn = {2752-6542},
    doi = {10.1093/pnasnexus/pgae231},
    url = {https://doi.org/10.1093/pnasnexus/pgae231},
    eprint = {https://academic.oup.com/pnasnexus/article-pdf/3/6/pgae231/58356613/pgae231.pdf},
}

@article{matz2024potential,
  title={The potential of generative AI for personalized persuasion at scale},
  author={Matz, Sandra C and Teeny, Jacob D and Vaid, Sumer S and Peters, Heinrich and Harari, Gabriella M and Cerf, Moran},
  journal={Scientific Reports},
  volume={14},
  number={1},
  pages={4692},
  year={2024},
  publisher={Nature Publishing Group UK London},
  doi={10.1038/s41598-024-53755-0},
  url={https://doi.org/10.1038/s41598-024-53755-0}
}

@article{hofstede1984hofstede,
  title={Hofstede's Culture Dimensions: An Independent Validation Using Rokeach's Value Survey},
  author={Hofstede, Geert and Bond, Michael Harris},
  journal={Journal of Cross-Cultural Psychology},
  volume={15},
  number={4},
  pages={417--433},
  year={1984},
  publisher={SAGE Publications},
  doi={10.1177/0022002184015004003},
  url={https://doi.org/10.1177/0022002184015004003}
}

@article{Shen2023LargeLM,
  title={Large Language Model Alignment: A Survey},
  author={Tianhao Shen and Renren Jin and Yufei Huang and Chuang Liu and Weilong Dong and Zishan Guo and Xinwei Wu and Yan Liu and Deyi Xiong},
  journal={ArXiv},
  year={2023},
  volume={abs/2309.15025},
  url={https://api.semanticscholar.org/CorpusID:262824801}
}

@techreport{radford2018improving,
  author = {Radford, Alec and Narasimhan, Karthik and Salimans, Tim and Sutskever, Ilya},
  institution = {OpenAI},
  title = {Improving language understanding by generative pre-training},
  year = 2018,
  url = {https://openai.com/research/language-unsupervised}
}

@misc{park2024rlhfheterogeneousfeedbackpersonalization,
      title={RLHF from Heterogeneous Feedback via Personalization and Preference Aggregation}, 
      author={Chanwoo Park and Mingyang Liu and Dingwen Kong and Kaiqing Zhang and Asuman Ozdaglar},
      year={2024},
      eprint={2405.00254},
      archivePrefix={arXiv},
      primaryClass={cs.AI},
      url={https://arxiv.org/abs/2405.00254}, 
}

@misc{sourati2025shrinkinglandscapelinguisticdiversity,
      title={The Shrinking Landscape of Linguistic Diversity in the Age of Large Language Models}, 
      author={Zhivar Sourati and Farzan Karimi-Malekabadi and Meltem Ozcan and Colin McDaniel and Alireza Ziabari and Jackson Trager and Ala Tak and Meng Chen and Fred Morstatter and Morteza Dehghani},
      year={2025},
      eprint={2502.11266},
      archivePrefix={arXiv},
      primaryClass={cs.CL},
      url={https://arxiv.org/abs/2502.11266}, 
}

@InProceedings{pmlr-v202-tennenholtz23a,
  title = 	 {Reinforcement Learning with History Dependent Dynamic Contexts},
  author =       {Tennenholtz, Guy and Merlis, Nadav and Shani, Lior and Mladenov, Martin and Boutilier, Craig},
  booktitle = 	 {Proceedings of the 40th International Conference on Machine Learning},
  pages = 	 {34011--34053},
  year = 	 {2023},
  editor = 	 {Krause, Andreas and Brunskill, Emma and Cho, Kyunghyun and Engelhardt, Barbara and Sabato, Sivan and Scarlett, Jonathan},
  volume = 	 {202},
  series = 	 {Proceedings of Machine Learning Research},
  month = 	 {23--29 Jul},
  publisher =    {PMLR},
  url = 	 {https://proceedings.mlr.press/v202/tennenholtz23a.html}
}

@inproceedings{
    rafailov2023direct,
    title={Direct Preference Optimization: Your Language Model is Secretly a Reward Model},
    author={Rafael Rafailov and Archit Sharma and Eric Mitchell and Christopher D Manning and Stefano Ermon and Chelsea Finn},
    booktitle={Thirty-seventh Conference on Neural Information Processing Systems},
    year={2023},
    url={https://arxiv.org/abs/2305.18290}
}

@inproceedings{10.5555/3540261.3542508,
author = {Lazaridou, Angeliki and Kuncoro, Adhiguna and Gribovskaya, Elena and Agrawal, Devang and Li\v{s}ka, Adam and Terzi, Tayfun and Gimenez, Mai and d'Autume, Cyprien de Masson and Kocisky, Tomas and Ruder, Sebastian and Yogatama, Dani and Cao, Kris and Young, Susannah and Blunsom, Phil},
title = {Mind the gap: assessing temporal generalization in neural language models},
year = {2021},
isbn = {9781713845393},
publisher = {Curran Associates Inc.},
address = {Red Hook, NY, USA},
booktitle = {Proceedings of the 35th International Conference on Neural Information Processing Systems},
articleno = {2247},
numpages = {16},
series = {NIPS '21}
}

@inproceedings{hashimoto_fairness_2018,
  title={Fairness Without Demographics in Repeated Loss Minimization},
  author={Hashimoto, Tatsunori B. and Srivastava, Megha and Namkoong, Hongseok and Liang, Percy},
  booktitle={International Conference on Machine Learning (ICML)},
  pages={1929--1938},
  year={2018},
  url={http://proceedings.mlr.press/v80/hashimoto18a.html}
}

@inproceedings{10.1145/3340531.3412152,
author = {Mansoury, Masoud and Abdollahpouri, Himan and Pechenizkiy, Mykola and Mobasher, Bamshad and Burke, Robin},
title = {Feedback Loop and Bias Amplification in Recommender Systems},
year = {2020},
isbn = {9781450368599},
publisher = {Association for Computing Machinery},
address = {New York, NY, USA},
url = {https://doi.org/10.1145/3340531.3412152},
doi = {10.1145/3340531.3412152},
booktitle = {Proceedings of the 29th ACM International Conference on Information \& Knowledge Management},
pages = {2145–2148},
numpages = {4},
location = {Virtual Event, Ireland},
series = {CIKM '20}
}

@article{radford2019language,
  title={Language models are unsupervised multitask learners},
  author={Radford, Alec and Wu, Jeffrey and Child, Rewon and Luan, David and Amodei, Dario and Sutskever, Ilya and others},
  journal={OpenAI blog},
  volume={1},
  number={8},
  pages={9},
  year={2019}
}

@inproceedings{
hendrycks2021measuring,
title={Measuring Massive Multitask Language Understanding},
author={Dan Hendrycks and Collin Burns and Steven Basart and Andy Zou and Mantas Mazeika and Dawn Song and Jacob Steinhardt},
booktitle={International Conference on Learning Representations},
year={2021},
url={https://openreview.net/forum?id=d7KBjmI3GmQ}
}

@article{pirolli1999information,
  title={Information foraging.},
  author={Pirolli, Peter and Card, Stuart},
  journal={Psychological review},
  volume={106},
  number={4},
  pages={643},
  year={1999},
  publisher={American Psychological Association}
}

@inproceedings{rose2004understanding,
  title={Understanding user goals in web search},
  author={Rose, Daniel E and Levinson, Danny},
  booktitle={Proceedings of the 13th international conference on World Wide Web},
  pages={13--19},
  year={2004}
}

@incollection{shani2010evaluating,
  title={Evaluating recommendation systems},
  author={Shani, Guy and Gunawardana, Asela},
  booktitle={Recommender systems handbook},
  pages={257--297},
  year={2010},
  publisher={Springer}
}

@inproceedings{ren2024representation,
  title={Representation learning with large language models for recommendation},
  author={Ren, Xubin and Wei, Wei and Xia, Lianghao and Su, Lixin and Cheng, Suqi and Wang, Junfeng and Yin, Dawei and Huang, Chao},
  booktitle={Proceedings of the ACM web conference 2024},
  pages={3464--3475},
  year={2024}
}

@inproceedings{covington2016deep,
  title={Deep neural networks for youtube recommendations},
  author={Covington, Paul and Adams, Jay and Sargin, Emre},
  booktitle={Proceedings of the 10th ACM conference on recommender systems},
  pages={191--198},
  year={2016}
}

@misc{hu2025reinforcestabilizingcriticfreepolicy,
      title={REINFORCE++: Stabilizing Critic-Free Policy Optimization with Global Advantage Normalization}, 
      author={Jian Hu and Jason Klein Liu and Haotian Xu and Wei Shen},
      year={2025},
      eprint={2501.03262},
      archivePrefix={arXiv},
      primaryClass={cs.CL},
      url={https://arxiv.org/abs/2501.03262}, 
}

@inproceedings{bu-etal-2025-personalized,
    title = "Personalized {LLM} Decoding via Contrasting Personal Preference",
    author = "Bu, Hyungjune  and
      Jung, ChanJoo  and
      Kang, Minjae  and
      Kim, Jaehyung",
    editor = "Christodoulopoulos, Christos  and
      Chakraborty, Tanmoy  and
      Rose, Carolyn  and
      Peng, Violet",
    booktitle = "Proceedings of the 2025 Conference on Empirical Methods in Natural Language Processing",
    month = nov,
    year = "2025",
    address = "Suzhou, China",
    publisher = "Association for Computational Linguistics",
    url = "https://aclanthology.org/2025.emnlp-main.1723/",
    doi = "10.18653/v1/2025.emnlp-main.1723",
    pages = "33958--33978",
    ISBN = "979-8-89176-332-6"
}

@inproceedings{yang2024rewards,
  title={Rewards-in-Context: Multi-objective Alignment of Foundation Models with Dynamic Preference Adjustment},
  author={Yang, Rui and Pan, Xiaoman and Luo, Feng and Qiu, Shuang and Zhong, Han and Yu, Dong and Chen, Jianshu},
  booktitle={International Conference on Machine Learning},
  pages={56276--56297},
  year={2024},
  organization={PMLR}
}

\newpage
\appendix
\onecolumn

\section{Practical Implementation}\label{appx:pratical-implementation}
To implement P-GRPO in practice, we must compute statistical moments over a growing sequence of rewards for each user preference $p$. A naive approach is to store the complete historic rewards $\mathcal{H}_p$ to recompute the mean and variance at every step, incurring linear memory complexity $O(N)$, which is prohibitive for large-scale distributed training. To address these constraints, we employ Welford’s online algorithm \citep{Welford1962}. This algorithm allows for the iterative update of running statistics with $O(1)$ memory complexity and high numerical stability. 

Algorithm~\ref{alg:p-grpo} presents the complete P-GRPO training procedure with online reward normalization. For each preference group $p$, we maintain three statistics: the count $n_p$, the running mean $\mu_p$, and the sum of squared differences $M_p$. When a new reward $r_i^p$ is observed for preference $p$, we update these statistics using Welford's algorithm to  compute the mean and variance without storing all historical rewards.

Note that during the early iterations, when $n_p$ is small, the variance estimate $\sigma_p^2 = M_p/(n_p - 1)$ can be unreliable due to insufficient samples.  We mitigate this with a warm-up phase: when $n_p$ falls below a minimum threshold $n_{\min}$, we record the rewards within the period to compute their mean and variance.

\begin{algorithm}[ht]
\caption{P-GRPO with Online Reward Normalization}
\label{alg:p-grpo}
\begin{algorithmic}[1]
\STATE \textbf{Input:} Policy $\policy$, reference policy $\refpolicy$, dataset $\mathcal{D}$, hyperparameters $c, \beta, \epsilon$
\STATE \textbf{Initialize:} For each preference $p$: $n_p \gets 0$, $\mu_p \gets 0$, $M_p \gets 0$
\FOR{epoch $= 1, 2, \ldots, T$}
    \FOR{batch $(\user, \prompt)$ sampled from $\mathcal{D}$}
        \STATE Sample group of completions $\{o_i\}_{i=1}^G \sim \policy(\cdot \mid \user, \prompt)$
        \STATE Compute rewards $\{r^p_i\}_{i=1}^G$ for each completion
        \FOR{each completion $i = 1, \ldots, G$}
            \STATE \textcolor{gray}{// Update running statistics using Welford's algorithm}
            \STATE $n_p \gets n_p + 1$
            \STATE $\delta_{\text{old}} \gets r^p_i - \mu_p$
            \STATE $\mu_p \gets \mu_p + \delta_{\text{old}} / n_p$ \quad \textcolor{gray}{// Update mean}
            \STATE $\delta_{\text{new}} \gets r^p_i - \mu_p$
            \STATE $M_p \gets M_p + \delta_{\text{old}} \cdot \delta_{\text{new}}$ \quad \textcolor{gray}{// Update variance}
            \STATE $\sigma_p \gets \sqrt{M_p / (n_p - 1)}$ if $n_p > 1$ else $1$
            \STATE // Compute personalized advantage
            \STATE $\tilde{A}_{i,t}^p \gets (r_i^p - \mu_p) / (\sigma_p + \epsilon)$ for all tokens $t \in o_i$
        \ENDFOR
        \STATE Compute importance ratios with \eqref{eq:importance-sampling}.
        \STATE Compute GRPO loss from \eqref{eq:L_GRPO} and \eqref{eq:token-wise-GRPO} and update $\theta$ by the optimizer.
    \ENDFOR
\ENDFOR
\STATE \textbf{Return:} Trained policy $\policy$
\end{algorithmic}
\end{algorithm}

\section{Example}
\label{apdx/linear_reward}
\begin{example}[Binary Reward Bias]
    Consider a binary reward setting where each completion receives a reward $R \in \{0, 1\}$ (fail/pass). Suppose the training data encompasses heterogeneous preference groups with different pass rates $\rho_p = \Pr(R = 1 \mid p)$. The reward mean and standard deviation for preference group $p$ are:
    \begin{align}
        \mu_p = \rho_p, \qquad \sigma_p = \sqrt{\rho_p(1 - \rho_p)}.
    \end{align}
    In standard GRPO, the intra-group normalization yields an advantage of approximately $\hat{A}(R{=}1) \approx \sqrt{(1-\rho_p)/\rho_p}$ for a passing completion and $\hat{A}(R{=}0) \approx -\sqrt{\rho_p/(1-\rho_p)}$ for a failing one. Taking the expectation over the Bernoulli reward, the expected absolute advantage for preference group $p$ is:
    \begin{equation}
        \mathbb{E}\bigl[|\hat{A}|\bigr] = \rho_p \sqrt{\tfrac{1-\rho_p}{\rho_p}} + (1-\rho_p)\sqrt{\tfrac{\rho_p}{1-\rho_p}} = 2\sqrt{\rho_p(1-\rho_p)},
    \end{equation}
    which is maximized at $\rho_p = 0.5$ and vanishes as $\rho_p \to 0$ or $\rho_p \to 1$. Because the expected absolute advantage governs the magnitude of the policy gradient update, GRPO's intra-group normalization introduces an implicit reweighting factor of $\sqrt{\rho_p(1-\rho_p)}$ across different preference groups: groups with moderate pass rates dominate the gradient, while groups with extreme pass rates are systematically suppressed.
\end{example}

    \paragraph{Numerical illustration.} Consider three preference groups with pass rates $\rho_1 = 0.9$ (easy), $\rho_2 = 0.1$ (hard), and $\rho_3 = 0.5$ (moderate). Under GRPO, the expected gradient signal strengths are:
    \begin{align}
        \mathbb{E}\bigl[|\hat{A}|\bigr]_{\rho_1=0.9} = 0.60, \qquad
        \mathbb{E}\bigl[|\hat{A}|\bigr]_{\rho_2=0.1} = 0.60, \qquad
        \mathbb{E}\bigl[|\hat{A}|\bigr]_{\rho_3=0.5} = 1.00.
    \end{align}
    The moderate preference group contributes stronger gradient signals than either extreme group, even when all three are equally represented in the training data. P-GRPO corrects this imbalance by normalizing each completion against its preference group's historical statistics $(\mu_p, \sigma_p)$, ensuring equitable gradient contributions across all preference groups regardless of their pass rates.
    
\section{Supplementary Related Works}
\subsection{Heterogeneous Rewards and Optimization Bias}

A central assumption in many preference-based learning algorithms is that observed rewards are samples from a single underlying distribution or are at least comparable across contexts \citep{park2024rlhfheterogeneousfeedbackpersonalization}. However, in practice, reward signals are often heterogeneous, arising from mixtures of latent user preferences, task types, or feedback mechanisms. In reinforcement learning, related challenges appear in contextual and multi-task RL, where policies must be learned under varying reward functions and noise profiles \citep{pmlr-v202-tennenholtz23a}.

Normalization and advantage estimation play a critical role in stabilizing policy gradient methods such as Proximal Policy Optimization (PPO) and GRPO. Prior work has shown that reward scaling and normalization can significantly affect learning dynamics, convergence, and bias \citep{yu2025dapo}. When rewards differ systematically in scale or variance across contexts, naive normalization can implicitly reweight different parts of the data, privileging high-variance or high-signal regimes while suppressing noisier or lower-variance ones.

Our analysis makes this effect explicit in the context of group-normalized preference optimization: when rewards reflect heterogeneous preference sensitivities, group-wise normalization induces a form of statistical shrinkage toward dominant or more discriminative preference modes. This results in biased learning that favors majority or “easier” preferences, even when minority preferences are equally important from a system design perspective.


\subsection{Fairness, Representation, and Minority Preferences}

The suppression of minority or under-represented signals connects our work to broader research on fairness and representation in machine learning systems. Prior work has documented how learning from imbalanced data can lead to biased models that underperform for minority groups, even when no explicit discriminatory intent is present \citep{hashimoto_fairness_2018}. In preference learning and recommender systems, aggregation of user feedback has been shown to amplify majority preferences and reduce diversity \citep{10.1145/3340531.3412152}.

In the context of LLM alignment, such effects are particularly concerning because preference optimization shapes the normative behavior of deployed systems. If certain user groups or interaction styles are systematically under-weighted during training, the resulting models may exhibit disparities in quality, relevance, or safety across populations.

Our approach can be interpreted as a debiasing mechanism for learning signals across heterogeneous preference groups. Instead of reweighting samples, P-GRPO ensures that the advantage for each completion is computed relative to its own preference-specific reward structure.

\subsection{Alignment and Preference Optimization}

Alignment research in LLMs seeks to ensure that model behavior reflects human values, intentions, and normative constraints. This includes SFT, RLHF \citep{ouyang2022training}, direct preference optimization (DPO) \citep{rafailov2023direct}, and group-based variants such as GRPO \citep{deepseek-math,yu2025dapo,gspo}. These methods have achieved substantial improvements in safety and helpfulness but are typically designed around homogeneous or implicitly aggregated preference signals.

Recent work has highlighted risks such as reward hacking \citep{macdiarmid2025naturalemergentmisalignmentreward} and goal misgeneralization \citep{pan2022the}, as well as sensitivity to distributional shift \citep{10.5555/3540261.3542508}. Our work complements this literature by identifying a structural limitation in group-normalized preference optimization under heterogeneous reward distributions and proposing a simple modification that preserves the benefits of GRPO while enabling equitable treatment of diverse preferences.

Together, these connections position P-GRPO as an optimization-level contribution to personalized alignment, addressing not how preferences are represented or elicited, but how they are integrated into learning mechanisms.

\section{Benchmark Details}
\subsection{Data Pre-processing for the MovieLens Benchmark.}
\label{apdx/movielens}

We utilize the MovieLens-1M dataset to construct a behavior modeling benchmark. The MovieLens-1M dataset comprises approximately one million ratings from 6,040 users on 3,706 movies, along with demographic information including age, gender, and occupation for each user. Our pre-processing pipeline transforms this dataset into a persona-conditioned sequential recommendation task, where the objective is to predict the next movie a user will watch given their user profile and viewing history.

User personas are derived directly from the demographic attributes provided in the dataset. Specifically, we map categorical age ranges (e.g., 18-24, 25-34), binary gender indicators (male, female), and occupation codes (21 categories ranging from academic/educator to writer) into natural language descriptions. For instance, a user with age group 25-34, gender male, and occupation programmer is represented as: ``A male at the age of 25-34 who works as programmer.'' To enable group-level policy optimization, we apply K-means clustering (k=10) on the one-hot encoded user features, thereby partitioning users into demographic clusters with similar characteristics.

The temporal structure of user interactions is preserved by sorting each user's interactions chronologically. For each user, we construct training instances using a sliding window approach over their viewing history. Given a sequence of the most recent movies watched, the task is formulated as a multiple-choice question where the model must select the next movie the user watched from among $N$ candidates: one positive example (the actual next movie) and $N-1$ negative examples (randomly sampled from the set of unwatched movies). This formulation enables direct application of preference optimization algorithms such as DPO and GRPO. We set $N=4$ in training and change a wide range of $N$ between 4 and 11 in the testing. We show the constructed prompts as follows:

\begin{promptbox}
You are a movie recommendation expert. Your task is to predict the next movie a user will watch based on their profile and viewing history.

\medskip
\noindent Your response must be a single, valid JSON object. The JSON object should contain one key: \texttt{"answer"}: The single letter (A, B, C, or D) corresponding to your choice.

\medskip
\noindent Example format: \texttt{\{"answer":"A"\}}

\medskip
\noindent \textbf{User Profile:}

\medskip
The user is \texttt{\{persona\}}

\medskip
\noindent \textbf{Viewing history:}

\medskip
\texttt{\{history\_sequence\}}

\medskip
\noindent \textbf{Question:}

\medskip
Based on this user's profile and viewing history, which of the following four movies would they most likely watch next?

\medskip
\texttt{\{movie\_choices\}}

\end{promptbox}

\subsection{Synthetic Review Data Generation}\label{apdx:SDG}
\begin{table}[h]
\centering
\setlength{\tabcolsep}{8pt}
\renewcommand{\arraystretch}{1.4}
\caption{Music genres used for data generation, their corresponding personas, and the associated disliked genres used to induce strongly negative reviews.}
\label{tbl:lists}
\small
\begin{tabular}{@{}lcp{9cm}@{}}
\toprule
\textbf{Genre} & \textbf{Disliked Genre} & \textbf{Persona Description} \\
\midrule
Jazz & EDM & A 65-year-old urban planner with a master's degree in public policy. Curious, introspective, and emotionally attuned, they value musical improvisation and reflective listening, and have a strong distaste for repetitive EDM music. \\
\addlinespace
Metal & Pop & Alex is a 40-year-old mechanic with a high school education. Introspective yet assertive and emotionally intense, Alex values authenticity and the cathartic energy of heavy music, and has a particular distaste for pop music. \\
\addlinespace
Pop & Metal & Zoe is a 13-year-old middle school student who is bright, energetic, and socially expressive. Curious and imaginative, she thrives on trends and emotional connection in music, but has a strong distaste for metal music. \\
\addlinespace
EDM & Folk & Mikey is a 23-year-old social media strategist with a bachelor's degree in marketing. Energetic and social, they are drawn to immersive electronic experiences and music festivals, and have a particular distaste for folk music. \\
\addlinespace
Reggaeton & Classical & Sofia is a 29-year-old marketing coordinator with a bachelor's degree in communications. Introverted but passionate about Latin music and dance, she uses reggaeton to relieve stress and connect with urban culture, and has a strong distaste for classical music. \\
\addlinespace
Classical & Reggaeton & Sandy, 34, holds a master's degree in musicology and works as a museum curator. Calm, introspective, and detail-oriented, she values structure and harmony in music and life, and has a particular distaste for reggaeton music. \\
\addlinespace
Folk & Jazz & Emma, 55, holds a master's degree in environmental studies and works in community outreach. Empathetic and reflective, she is drawn to folk music's storytelling and acoustic simplicity, and has a particular distaste for jazz music. \\
\bottomrule
\end{tabular}
\end{table}
Here we provide detailed information for the synthetic review data generation process. 

We considered seven music genres for data generation, chosen to span a diverse range of styles and to minimize overlap between the resulting groups or clusters. For each genre, we defined a corresponding persona, which is used to guide the tone and sentiment of the generated reviews and to encourage stylistic differentiation across groups. Reviews are generated conditioned on both the song’s genre and the reviewer’s persona. When the song genre matches the reviewer’s persona, the review is positive; when it does not, the review is negative. Additionally, each persona is associated with a specific disliked genre. Reviews of songs from this genre are generated with a strongly negative tone. Table \ref{tbl:lists} summarizes the genres, their corresponding personas (and their associated disliked genres).

We first consider the genres listed in Table~\ref{tbl:lists} and generate a dataset of $200$ songs per genre, along with selected song-level characteristics. This dataset was generated using the Qwen3-32B model with the following prompt.

\begin{promptbox}
Generate \texttt{\{batch\_size\}} realistic \texttt{\{genre\}} songs with diverse titles, artists, and musical characteristics.

\medskip
\noindent For each song, provide:
\begin{itemize}
    \item track\_name: A creative, realistic song title
    \item artist\_name: A realistic artist or band name
    \item danceability: How suitable for dancing (0.0 to 1.0, where 1.0 is most danceable)
    \item energy: Intensity and activity level (0.0 to 1.0, where 1.0 is highest energy)
    \item loudness: Overall loudness in decibels (typically -60 to 0 dB)
    \item tempo: Beats per minute (typically 40 to 200 BPM)
    \item duration\_ms: Song length in milliseconds (typically 120000 to 600000 ms)
\end{itemize}
\medskip
\noindent Make the numerical values realistic for \texttt{\{genre\}} music:
\begin{itemize}
\item Consider typical \texttt{\{genre\}} characteristics when assigning values
\item Ensure values are coherent (e.g., high energy songs often have higher loudness and tempo)
\item Use your knowledge of \texttt{\{genre\}} music to provide appropriate ranges
\end{itemize}

\medskip
\noindent Make the songs diverse and realistic. Use actual \texttt{\{genre\}} naming conventions and styles.
Generate exactly \texttt{\{batch\_size\}} songs.
\end{promptbox}

Once the song metadata is generated, we use it for review generation. We again employ the Qwen3-32B model, using two separate prompts: one prompt generates the song review, and the other generates an explanation of the song. The generated reviews serve as the output data samples, while the generated explanations are used as the input data (prompts) for training in the P-GRPO framework. We designed the prompts to condition the reviewer’s attitude on the relationship between the persona and the song genre. When the persona and genre match, the prompt elicits a positive review; when they do not match, it produces a non-positive review (negative or strongly negative).

\begin{promptbox}
Generate reviews:

\medskip
\noindent You are going to write a \texttt{\{level\}} review of the following song: \texttt{\{title\}} that is performed with the artist \texttt{\{artist\}} and has the genre \texttt{\{genre\}}.

\medskip
\noindent But you must first infer the persona's unique writing style.

\medskip
\noindent Follow these steps carefully for the following persona: \texttt{\{persona\}}

\medskip
\noindent Firstly Infer Persona Style
\begin{enumerate}
\item You are given a persona (e.g., a famous author, journalist, celebrity, philosopher, or fictional character).
\item Reflect on this persona's likely writing or speaking style. Consider:
  \begin{itemize}
  \item Tone: formal, casual, humorous, dramatic, sarcastic, poetic, etc.
  \item Vocabulary: simple, complex, archaic, modern, flowery, technical
  \item Sentence structure: long and elaborate, short and punchy, rhythmic
  \item Emotional expressiveness: reserved, passionate, enthusiastic
  \item Signature phrases, metaphors, or patterns
  \end{itemize}
\item Write a concise summary describing this inferred style.
\end{enumerate}

\medskip
\noindent Next Analyze the Song
\begin{itemize}
\item Consider key aspects of the song: melody, lyrics, rhythm, instrumentation, emotional impact, or vocals.
\item Think about which aspects would resonate most with this persona.
\end{itemize}

\medskip
\noindent In addition, let's Generate Review Using Persona Style
\begin{itemize}
\item Using the inferred style, write a {level} review of the song.
\item Make sure the tone, vocabulary, and sentence structure match the persona's style.
\item Explain why the song is exceptional, highlighting details the persona would naturally notice.
\item Include enthusiasm and emotional impact consistent with the persona's voice.
\end{itemize}

\medskip
\noindent and Finally Output the full song review in that style.

\medskip
\noindent Please also use less hyphens.

\medskip
\noindent Respond in JSON format with the following fields:
\begin{itemize}
\item "review": Your generated music review (written entirely in the persona's authentic voice)
\item "persona\_description": Brief one-line description of the persona
\item "explanation": Specific examples from your review showing how language, references, and perspectives unmistakably reveal this persona
\item "rating": The star rating (1-5) you chose for this song
\end{itemize}
\end{promptbox}

\begin{promptbox}
Based on your knowledge about music, provide a brief description of the following song:

\medskip
\noindent Title: \texttt{\{title\}}

\medskip
\noindent Artist: \texttt{\{artist\}}

\medskip
\noindent Characteristics: {\{characteristics\}}

\medskip
\noindent Please provide a concise 2-3 sentence description that captures the essence, style, and notable features of this song based on your knowledge of it. If you don't know the specific song, provide a general description based on the artist's typical style and the given characteristics.

\medskip
\noindent Respond with just the description text, no additional formatting.
\end{promptbox}

To maintain a balance between positive and negative reviews for each persona, we first generate 200 reviews for songs whose genres match the persona, and then sample 200 additional songs from the remaining genres to generate non-positive reviews. This yields 400 samples per persona, resulting in seven distinct clusters, each corresponding to a single persona. This design ensured systematic variation in review valence as a function of genre alignment while simultaneously inducing variation in linguistic style across personas.

\subsection{Data Pre-processing for Goodreads Book Reviews}
\label{apdx/goodreads}
We utilize the Goodreads Books dataset to construct a benchmark for personalized review generation. The dataset comprises rich book metadata and user-authored reviews collected from the Goodreads platform. Our pre-processing pipeline transforms this dataset into a rating-conditioned review generation task, where the goal is to generate a personalized book review given structured book information and the target rating.

Book metadata is extracted from the dataset and includes key attributes such as title, author, description, genres, average rating, number of pages, publication year, and publisher. For genre information, we extract the top 5 relevant genres from the popular shelves while filtering out generic categories. Review texts are cleaned to remove spoiler tags and excessive whitespace while preserving the original content. Additionally, we apply a minimum review length threshold of 50 characters to exclude uninformative or overly brief reviews.

A key characteristic of this dataset is the absence of explicit user demographic information. To accommodate this limitation within our personalization framework, we use the review rating (1–5 stars) as a proxy for preference cluster identifiers, thereby grouping users according to sentiment alignment. This design enables preference-based clustering, where each rating level corresponds to a distinct user preference group with differing sentiment orientations toward books. 

We show the constructed prompts as follows:
\begin{promptbox}
Based on the book information below, write a detailed \texttt{\{rating\}}-star review:

\medskip
\noindent Title: \texttt{\{title\}}

\medskip
\noindent Author: \texttt{\{author\}}

\medskip
\noindent Average Rating: \texttt{\{average\_rating\}}

\medskip
\noindent Pages: \texttt{\{num\_pages\}}

\medskip
\noindent Publication Year: \texttt{\{publication\_year\}}

\medskip
\noindent Publisher: \texttt{\{publisher\}}

\medskip
\noindent Genres: \texttt{\{genres\}}

\medskip
\noindent Description: \texttt{\{description\}}
\end{promptbox}

\subsection{Data Pre-processing for KG-Rec Music Benchmark}
\label{apdx/kgrec}

The KG-Rec dataset integrates implicit user-item interactions with rich textual descriptions and semantic tags sourced from external music databases. User listening sequences are constructed from implicit feedback data, wherein each user's interaction history represents their chronological sequence of listened songs. Due to the sparsity of explicit demographic attributes in this dataset, we derive user personas through behavioral clustering rather than demographic features. Specifically, we generate textual representations for each user's listening sequence, followed by K-means clustering (k=10) on the resulting feature vectors. This approach groups users with similar interests into preference clusters that can be further used in our P-GRPO algorithm.

For each item (music), we leverage two types of metadata provided by the knowledge graph: (1) textual descriptions that summarize the song's musical characteristics, lyrical themes, and contextual information, and (2) semantic tags representing categorical attributes such as genre, mood, and instrumentation. These metadata fields are concatenated into a unified item representation that serves as both input context and output reference for the preference learning task.

The prediction task is formulated as a generative completion problem. Given a user's listening history of recent songs (including item IDs and descriptions), the model is prompted to generate a textual description of the next song the user will listen to. The ground truth reference consists of the actual next song's description and tags, enabling the application of sequence-level preference optimization methods. This formulation differs from the classification-based approach used in MovieLens, thereby providing complementary insights into the effectiveness of \textit{P-GRPO} across diverse task structures.

\begin{promptbox}
You are an expert music recommender.

\medskip
\noindent \textbf{User Listening History:}

\medskip
\texttt{\{history\_sequence\}}

\medskip
\noindent Based on the user's listening history, please describe the next song the user is most likely to listen to.
\end{promptbox}

\section{Additional Experiment Results}\label{appendix:additional-experiment-results}

\subsection{Detailed Results for Language Generation}\label{apdx/add_exp}
\begin{table*}[!ht]
    \centering
    \caption{Test-set generation performance on KGRec-Music and Goodreads Books datasets.}
    \label{tab:appendix/result/main}
    \small
    \resizebox{\textwidth}{!}{
        \begin{tabular}{@{}lll cccc@{}}
            \toprule
            \textbf{Dataset} & \textbf{Model} & \textbf{Method} & Rouge-1 & Rouge-2 & Rouge-L & Cos. Sim \\
            \midrule
                \multirow{6}{*}{Synthetic Data}
                & \multirow{3}{*}{Gemma-2B}
                & GDPO & $0.5297 \pm 0.1275$ & $0.1329 \pm 0.0577$ & $0.2907 \pm 0.0858$ & $0.7050 \pm 0.1675$ \\
                & & GRPO & $0.6042 \pm 0.1276$ & $0.1775 \pm 0.0703$ & $0.3706 \pm 0.0948$ & $0.7090 \pm 0.1662$ \\
                & & P-GRPO & $\mathbf{0.6133 \pm 0.1286}$ & $\mathbf{0.1861 \pm 0.0717}$ & $\mathbf{0.3798 \pm 0.0970}$ & $\mathbf{0.7154 \pm 0.1683}$ \\
                \cmidrule{2-7}
                & \multirow{3}{*}{Qwen3-8B}
                & GDPO & $0.5663 \pm 0.1422$ & $0.1374 \pm 0.0735$ & $0.3359 \pm 0.0986$ & $0.5246 \pm 0.1677$ \\       
                & & GRPO & $0.6075 \pm 0.1430$ & $0.1713 \pm 0.0797$ & $0.3624 \pm 0.1024$ & $0.5814 \pm 0.1764$ \\
                & & P-GRPO & $\mathbf{0.6267 \pm 0.1370}$ & $\mathbf{0.1853 \pm 0.0778}$ & $\mathbf{0.3758 \pm 0.1011}$ & $\mathbf{0.6150 \pm 0.1793}$ \\
            \midrule
                \multirow{6}{*}{KGRec}
                & \multirow{3}{*}{Gemma-2B}
                & GDPO & $0.2513 \pm 0.0966$ & $0.0226 \pm 0.0220$ & $0.1455 \pm 0.0659$ & $0.3088 \pm 0.0833$ \\
                & & GRPO & $0.5603 \pm 0.0767$ & $0.1058 \pm 0.0401$ & $0.2832 \pm 0.0867$ & $0.3069 \pm 0.0943$ \\
                & & P-GRPO & $\mathbf{0.5618 \pm 0.0717}$ & $\mathbf{0.1067 \pm 0.0397}$ & $\mathbf{0.2843 \pm 0.0859}$ & $\mathbf{0.3130 \pm 0.0943}$ \\
                \cmidrule{2-7}
                & \multirow{3}{*}{Qwen3-8B} 
                & GDPO & $0.3845 \pm 0.0762$ & $0.0473 \pm 0.0279$ & $0.2419 \pm 0.0703$ & $\mathbf{0.3177 \pm 0.0873}$\\
                & & GRPO & $0.4340 \pm 0.0908$ & $0.0790 \pm 0.0407$ & $\mathbf{0.2848 \pm 0.0817}$ & $0.2905 \pm 0.0872$ \\
                & & P-GRPO & $\mathbf{0.4649 \pm 0.1067}$ & $\mathbf{0.0856 \pm 0.0439}$ & $0.2840 \pm 0.0834$ & $0.2907 \pm 0.0906$ \\
            \midrule
                \multirow{6}{*}{Goodreads}
                & \multirow{3}{*}{Gemma-2B}
                & GDPO & $0.4008 \pm 0.1119$ & $0.0600 \pm 0.0420$ & $0.2254 \pm 0.0984$ & $0.5288 \pm 0.1469$ \\
                & & GRPO & $0.5526 \pm 0.1053$ & $0.1060 \pm 0.0551$ & $\mathbf{0.3255 \pm 0.1160}$ & $0.5328 \pm 0.1502$ \\
                & & P-GRPO & $\mathbf{0.5534 \pm 0.0934}$ & $\mathbf{0.1181 \pm 0.1053}$ & $0.3204 \pm 0.1255$ & $\mathbf{0.5415 \pm 0.1444}$ \\
                \cmidrule{2-7}
                & \multirow{3}{*}{Qwen3-8B}
                & GDPO & $0.4987 \pm 0.1010$ & $0.0781 \pm 0.0465$ & $0.3003 \pm 0.1097$ & $ 0.4520 \pm 0.1379$ \\
                & & GRPO & $0.6076 \pm 0.1054$ & $0.1374 \pm 0.0675$ & $\mathbf{0.3596 \pm 0.1295}$ & $0.5266 \pm 0.1557$ \\
                & & P-GRPO & $\mathbf{0.6138 \pm 0.1042}$ & $\mathbf{0.1383 \pm 0.0659}$ & $0.3578 \pm 0.1294$ & $\mathbf{0.5296 \pm 0.1542}$ \\
            \bottomrule
        \end{tabular}
    }
\end{table*}
We report the comprehensive test set performance across three generative tasks in Table~\ref{tab:appendix/result/main}. All metrics are reported as mean $\pm$ standard deviation over the test set.

\subsection{Preservation of General Capabilities}
\label{apdx/mmlu}

To demonstrate that our Personalized GRPO algorithm does not harm the original general capabilities of the language models, we evaluate the models on the Massive Multitask Language Understanding (MMLU) benchmark \citep{hendrycks2021measuring}. MMLU is a comprehensive benchmark consisting of 57 diverse tasks spanning STEM, humanities, social sciences, and other areas, designed to measure a model's broad knowledge and reasoning capabilities. 

We compare the MMLU performance of models before and after applying P-GRPO training on our personalized behavior modeling tasks in Table~\ref{tab:result/mmlu}. For Qwen3-8B, the results demonstrate that P-GRPO maintains nearly identical performance to the pre-trained model, with accuracy differences within $\pm0.06$ across all the personalization tasks. For Gemma-2B, we observe slightly larger but still modest decreases ranging from $-0.35\%$ to $-0.60\%$, which remain within acceptable bounds for specialized post-training. These results confirm that P-GRPO effectively adapts models to user-specific preferences while preserving their fundamental language understanding and reasoning capabilities.

\begin{table}[ht]
    \centering
    \caption{MMLU benchmark performance comparing pre-trained models with their P-GRPO fine-tuned counterparts across different personalization tasks. Accuracy scores and deltas from base models are reported for Qwen3-8B and Gemma-2B..}
    \label{tab:result/mmlu}
    \small
    \begin{tabular}{@{}lcc@{}}
        \toprule
            \textbf{Model} & \textbf{Accuracy} & \textbf{$\Delta$ from Base} \\
         \midrule
            Qwen3-8B (Pre-trained)           & 71.29\% & -- \\
            \quad + P-GRPO on MovieLens      & 71.25\% & -0.04\% \\
            \quad + P-GRPO on Synthetic Data & 71.23\% & -0.06\% \\
            \quad + P-GRPO on KGRec          & 71.31\% & +0.02\% \\
            \quad + P-GRPO on Goodreads      & 71.29\% & 0.00\% \\
         \midrule
            Gemma-2B (Pre-trained)           & 57.27\% & -- \\
            \quad + P-GRPO on MovieLens      & 56.67\% & -0.60\% \\
            \quad + P-GRPO on Synthetic Data & 57.26\% & -0.01\% \\
            \quad + P-GRPO on KGRec          & 56.89\% & -0.38\% \\
            \quad + P-GRPO on Goodreads      & 56.92\% & -0.35\% \\
        \bottomrule
    \end{tabular}
\end{table}

\subsection{Robustness Analysis}\label{apdx/robustness}

We complement the clustering-quality ablations in Section~\ref{sec:exp} with three controlled studies that stress-test P-GRPO's reliance on cluster assignments: robustness to label noise, sensitivity to cluster granularity, and sensitivity to cluster geometry. Throughout, we report the cross-cluster disparity $\Delta$, defined as the accuracy gap between the best- and worst-performing clusters (lower is more equitable).

\subsubsection{Robustness to Cluster Label Noise}\label{apdx/noise}
To assess robustness to misclassified clusters, we sweep the cluster-label flip rate $\eta \in \{0, 0.25, 0.5, 0.75, 1.0\}$ on MovieLens-1M with Qwen3-8B, where at rate $\eta$ a sample's cluster ID is replaced by a uniformly random one (e.g., $\eta=0.5$ means half of the labels are random). As shown in Table~\ref{tab:robust/noise}, accuracy degrades gracefully as noise increases and the gain over vanilla GRPO vanishes at $\eta=1.0$, exactly as expected since fully random labels reduce P-GRPO to a global running-mean baseline. The disparity $\Delta$ likewise grows with noise and regresses toward the baseline level, confirming that accurate clustering is what drives equitable performance.

\begin{table}[ht]
    \centering
    \caption{Robustness to cluster label noise on MovieLens-1M (Qwen3-8B). $\eta$ is the fraction of cluster labels replaced by random assignments.}
    \label{tab:robust/noise}
    \small
    \begin{tabular}{@{}lccc@{}}
        \toprule
        \textbf{Method} & \textbf{Noise $\eta$} & \textbf{Accuracy} & \textbf{Cluster $\Delta$} \\
        \midrule
        Vanilla GRPO & N/A  & 0.7190 & 0.1247 \\
        P-GRPO       & 0.00 & 0.7566 & 0.1130 \\
        P-GRPO       & 0.25 & 0.7516 & 0.1170 \\
        P-GRPO       & 0.50 & 0.7428 & 0.1200 \\
        P-GRPO       & 0.75 & 0.7393 & 0.1870 \\
        P-GRPO       & 1.00 & 0.7203 & 0.1840 \\
        \bottomrule
    \end{tabular}
\end{table}

\subsubsection{Cluster Granularity}\label{apdx/granularity}
To characterize how performance varies with the number of clusters, we sweep $k \in \{1,5,10,15,20,25,30,50\}$ on MovieLens-1M with Qwen3-8B (Table~\ref{tab:robust/ksweep}). Aggregate accuracy improves with finer granularity, while the disparity $\Delta$ reaches its minimum (most equitable) at $k=20$. Notably, an independent Silhouette analysis identifies $k=20$ as the optimal number of clusters for this dataset, so our RL results align with this purely geometric heuristic. At very large $k$ ($k \ge 30$), disparity rises again, consistent with sample starvation in overly fragmented clusters.

\begin{table}[ht]
    \centering
    \caption{Effect of cluster granularity $k$ on MovieLens-1M (Qwen3-8B).}
    \label{tab:robust/ksweep}
    \small
    \begin{tabular}{@{}lcc@{}}
        \toprule
        \textbf{$k$} & \textbf{Accuracy} & \textbf{Cluster $\Delta$} \\
        \midrule
        1   & 0.7299 & 0.101 \\
        5   & 0.7522 & 0.085 \\
        10  & 0.7566 & 0.113 \\
        15  & 0.7520 & 0.102 \\
        20  & 0.7566 & 0.069 \\
        25  & 0.7612 & 0.094 \\
        30  & 0.7467 & 0.169 \\
        50  & 0.7409 & 0.269 \\
        \bottomrule
    \end{tabular}
\end{table}

\subsubsection{Cluster Geometry}\label{apdx/geometry}
Beyond label noise, we evaluate sensitivity to cluster geometry using the synthetic text-generation setting. We construct a graded sequence of persona sets by initializing seven random personas and iteratively updating them to maximize the distance between the centroids of their corresponding reviews in embedding space (using Qwen3-32B for generation and Qwen-Embedder-8B for embeddings), capturing a snapshot every five iterations over 50 iterations to yield 10 distinct sets. The ``grade'' indexes the concentration of clusters: lower grades correspond to dispersed, distinct personas and higher grades to highly concentrated, overlapping personas. As clusters become more concentrated (Table~\ref{tab:robust/geometry}), the within-cluster variance $\sigma_p$ shrinks and, consistent with Corollary~\ref{cor:decomposition}, the per-cluster rescaling collapses toward vanilla GRPO. P-GRPO thus automatically and safely reduces its personalization signal when distinct preference groups cannot be cleanly separated.

\begin{table}[ht]
    \centering
    \caption{Effect of cluster geometry on the synthetic dataset. Higher grade indicates more concentrated, overlapping clusters.}
    \label{tab:robust/geometry}
    \small
    \begin{tabular}{@{}lcccc@{}}
        \toprule
        \textbf{Grade} & \textbf{ROUGE-1} & \textbf{ROUGE-2} & \textbf{ROUGE-L} & \textbf{Reward} \\
        \midrule
        1  & 0.5141 & 0.0746 & 0.0680 & 0.6567 \\
        2  & 0.5147 & 0.0773 & 0.0695 & 0.6615 \\
        3  & 0.4897 & 0.0653 & 0.0693 & 0.6243 \\
        4  & 0.4865 & 0.0638 & 0.0723 & 0.6226 \\
        5  & 0.4812 & 0.0665 & 0.0780 & 0.6257 \\
        6  & 0.4837 & 0.0596 & 0.0742 & 0.6175 \\
        7  & 0.4758 & 0.0562 & 0.0717 & 0.6037 \\
        8  & 0.4809 & 0.0633 & 0.0743 & 0.6185 \\
        9  & 0.4827 & 0.0626 & 0.0792 & 0.6245 \\
        10 & 0.4778 & 0.0595 & 0.0755 & 0.6128 \\
        \bottomrule
    \end{tabular}
\end{table}

\subsection{Interaction with PPO Clipping}\label{apdx/clipping}
We expand on the remark following Corollary~\ref{cor:decomposition}. PPO clipping operates on the importance ratio $\rho_{i,t}$ rather than on the advantage magnitude, so the rescaling factor $\sigma_G/\sigma_p$ acts as an informative cross-cluster reweighting with three regimes:
\begin{enumerate}[leftmargin=*]
    \item $\sigma_p \ll \sigma_G$: updates are amplified, driving $\rho_{i,t}$ away from $1$ more rapidly. Clipping activates earlier, acting as an adaptive trust region rather than permitting uncontrolled updates. This is precisely the mechanism that counteracts the suppression of minority/extreme clusters.
    \item $\sigma_p \gg \sigma_G$: updates are suppressed, implicitly down-weighting noisier clusters whose rewards carry less information per example, yielding a more conservative update.
    \item $\sigma_p \sim \sigma_G$: the rescaling is negligible and P-GRPO recovers standard GRPO.
\end{enumerate}
The bias-correction term $(\mu_G - \mu_p)/\sigma_p$ acts as a cluster-level progress signal: it shifts advantages upward when a generation group outperforms its cluster's historical average and downward otherwise, providing an appropriate signal for that preference group rather than forcing advantages to sum to zero within each generation group. A simultaneous collapse of this term across all clusters corresponds to the personalization-collapse regime. To address blow-up when $\sigma_p$ is small early in training, the warm-up threshold $n_{\min}$ (Appendix~\ref{appx:pratical-implementation}) falls back to batch statistics until a stable variance estimate forms. In practice, the small learning rates and KL penalties of RLHF naturally constrain single-step policy shifts: we monitored $\rho_{i,t}$ throughout training and found it concentrated within $[0.95, 1.05]$ across all steps, confirming that the clipping operator almost never activates and that the rescaling does not silently nullify gradient updates.

\subsection{Robustness of the LLM-as-Judge Evaluation}\label{apdx/judge-robustness}
To validate the win-rate evaluation in Section~\ref{sec:exp}, we assess both inter-judge agreement and prompt sensitivity. We re-run the pairwise evaluation with a second, independent judge (Gemma-2-27B) in addition to GPT-OSS-120B. The two judges corroborate each other at moderate agreement, with Cohen's $\kappa = 0.52$ on KGRec and $\kappa = 0.46$ on Goodreads, indicating consistent agreement on the overall ranking of P-GRPO above GRPO with example-level disagreement expected for this subjective task. To assess prompt sensitivity, we re-run the evaluation across three prompt variants (a minimal version, a more structured and descriptive version, and a version with strict criteria). As shown in Table~\ref{tab:judge/prompt}, the P-GRPO-vs-GRPO win rate varies by only $0.045$ on KGRec and $0.077$ on Goodreads, and remains above $50\%$ in all configurations.

\begin{table}[ht]
    \centering
    \caption{LLM-as-judge win rates (\%) of P-GRPO over GRPO across two judges and three prompt variants.}
    \label{tab:judge/prompt}
    \small
    \begin{tabular}{@{}llccc@{}}
        \toprule
        \textbf{Data} & \textbf{Judge} & \textbf{Minimal} & \textbf{Structured} & \textbf{Criteria} \\
        \midrule
        \multirow{2}{*}{KGRec}    & GPT-OSS-120B & 51.8 & 54.6 & 50.1 \\
                                  & Gemma-2-27B  & 52.6 & 55.4 & 56.0 \\
        \midrule
        \multirow{2}{*}{Goodreads}& GPT-OSS-120B & 70.0 & 69.3 & 68.3 \\
                                  & Gemma-2-27B  & 75.0 & 69.7 & 67.3 \\
        \bottomrule
    \end{tabular}
\end{table}

\end{document}